\documentclass{article}

\usepackage{todonotes}

\newcommand{\hide}[1]{}

\usepackage{amsmath}
\usepackage{amssymb}
\usepackage{amsfonts}
\usepackage{amsthm}

\newtheorem{mydefinition}{Definition}
\newtheorem{proposition}[mydefinition]{Proposition}
\newtheorem{theorem}[mydefinition]{Theorem}
\newtheorem{remark}[mydefinition]{Remark}
\newtheorem{lemma}[mydefinition]{Lemma}

\renewcommand{\exp}[1]{\operatorname{exp}\left(#1\right)}
\newcommand{\diam}{\operatorname{diam}}
\newcommand{\E}[1]{\mathbb{E}\left[#1\right]}
\renewcommand{\P}[1]{\mathbb{P}\left(#1\right)}
\newcommand{\1}[1]{\mathbb{I}\left\{#1\right\}}
\newcommand{\R}{\mathbb{R}}
\newcommand{\ceil}[1]{\left\lceil#1\right\rceil}

\usepackage{graphicx} 
\usepackage{subfigure} 

\usepackage{tikz}
\usetikzlibrary{calc,intersections}

\usepackage{natbib}

\usepackage{algorithm}
\usepackage{algorithmic}

\usepackage{hyperref}

\usepackage[accepted]{icml2013}

\icmltitlerunning{Consistency of Online Random Forests}

\begin{document}

\twocolumn[
\icmltitle{Consistency of Online Random Forests}

\icmlauthor{Misha Denil}{mdenil@cs.ubc.ca}
\icmlauthor{David Matheson}{davidm@cs.ubc.ca}
\icmlauthor{Nando de Freitas}{nando@cs.ubc.ca}
\icmladdress{University of British Columbia}

\icmlkeywords{machine learning, ICML}

\vskip 0.3in
]

\begin{abstract}
  As a testament to their success, the theory of random forests has
  long been outpaced by their application in practice.  In this paper,
  we take a step towards narrowing this gap by providing a consistency
  result for online random forests.
\end{abstract}

\section{Introduction}
\label{sec:introduction}

Random forests are a class of ensemble method whose base learners are
a collection of randomized tree predictors, which are combined through
averaging.  The original random forests framework described in
\citet{Breiman2001:random_forests} has been extremely influential
\cite{Svetnik2003, Prasad2006, Cutler2007, Shotton2011,
  criminisi2011decision}.

Despite their extensive use in practical settings, very little is
known about the mathematical properties of these algorithms.  A recent
paper by one of the leading theoretical experts states that
\begin{quote}
  Despite growing interest and practical use, there has been little
  exploration of the statistical properties of random forests, and
  little is known about the mathematical forces driving the
  algorithm \cite{Biau2012}.
\end{quote}

Theoretical work in this area typically focuses on stylized versions
of the random forests algorithms used in practice.  For example,
\citet{biau08} prove the consistency of a variety of ensemble methods
built by averaging base classifiers.  Two of the models they study are
direct simplifications of the forest growing algorithms used in
practice; the others are stylized neighbourhood averaging rules, which
can be viewed as simplifications of random forests through the lens of
\citet{Lin2002}.  An even further simplified version of random forests in one
dimension is studied in \citet{genuer2010risk, genuer2012variance}.

In this paper we make further steps towards narrowing the gap between
theory and practice.  In particular, we present what is, to the best of
our knowledge, the first consistency result for online random forests.

%


\section{Related Work}

Different variants of random forests are distinguished by the methods
they use for growing the trees.  The model described in
\citet{Breiman2001:random_forests} builds each tree on a bootstrapped
sample of the training set using the CART
methodology~\cite{breiman84}.  The optimization in each leaf that
searches for the optimal split point is restricted to a random
selection of features, or linear combinations of features.

The framework of \citet{criminisi2011decision} operates slightly
differently.  Instead of choosing only features at random, this
framework chooses entire decisions (i.e.\ both a feature or
combination of features and a threshold together) at random and
optimizes only over this set.
Unlike the work
of \citet{Breiman2001:random_forests}, this framework chooses not to
include bagging, preferring instead to train each tree on the entire
data set and introduce randomness only in the splitting process.  The
authors argue that without bagging their model obtains max-margin
properties.

In addition to the frameworks mentioned above, many practitioners
introduce their own variations on the basic random forests algorithm,
tailored to their specific problem domain.  A variant from
\citet{Bosch} is especially similar to the technique we use in this
paper: When growing a tree the authors randomly select one third of
the training data to determine the structure of the tree and use the
remaining two thirds to fit the leaf estimators.  However, the authors
consider this only as a technique for introducing randomness into the
trees, whereas in our model the partitioning of data plays a central
role in consistency.

In addition to these offline methods, several researchers have focused
on building online versions of random forests.  Online models are
attractive because they do not require that the entire training set be
accessible at once.  These models are appropriate for streaming
settings where training data is generated over time and should be
incorporated into the model as quickly as possible.  Several variants
of online decision tree models are present in the MOA system of
\citet{Bifet2010}.

The primary difficulty with building online decision trees is their
recursive structure.  Data encountered once a split has been made cannot
be used to correct earlier decisions.  A notable approach to this
problem is the Hoeffding tree \cite{domingos2000mining} algorithm,
which works by maintaining several candidate splits in each leaf.  The
quality of each split is estimated online as data arrive in the leaf,
but since the entire training set is not available these quality
measures are only estimates.  The Hoeffding bound is employed in each
leaf to control the amount of data which must be collected to ensure
that the split chosen on the basis of these estimates is the true best
split with high probability.  \citet{domingos2000mining} prove that
under reasonable assumptions the online Hoeffding tree converges to
the offline tree with high probability.  The Hoeffding tree algorithm
is implemented in the system of \citet{Bifet2010}.

Alternative methods for controlling tree growth in an online setting
have also been explored.  \citet{saffari2009line} use the online
bagging technique of \citet{Oza2001} and control leaf splitting using
two parameters in their online random forest.  One parameter
specifies the minimum number of data points which must be seen in a
leaf before it can be split, and another specifies a minimum quality
threshold that the best split in a leaf must reach.  This is similar
in flavor to the technique used by Hoeffding trees, but trades
theoretical guarantees for more interpretable parameters.

One active avenue of research in online random forests involves
tracking non-stationary distributions, also known as concept drift.
Many of the online techniques incorporate features designed for this
problem \cite{Gama:2005:LDT:1066677.1066809, Abdulsalam2008,
  saffari2009line, Bifet2009, Bifet2012}.  However, tracking of
non-stationarity is beyond the scope of this paper.

The most well known theoretical result for random forests is that of
\citet{Breiman2001:random_forests}, which gives an upper bound on the
generalization error of the forest in terms of the correlation and
strength of trees.  Following \citet{Breiman2001:random_forests}, an
interpretation of random forests as an adaptive neighborhood weighting
scheme was published by \citet{Lin2002}.  This was followed by the
first consistency result in this area from \citet{Breiman2004}, who
proves consistency of a simplified model of the random forests used in
practice.  In the context of quantile regression the consistency of a
certain model of random forests has been shown by
\citet{Meinshausen2006}.  A model of random forests for survival
analysis was shown to be consistent in \citet{Ishwaran2010}.

Significant recent work in this direction comes from \citet{biau08}
who prove the consistency of a variety of ensemble methods built by
averaging base classifiers, as is done in random forests.  A key
feature of the consistency of the tree construction algorithms they
present is a proposition that states that if the base classifier is
consistent then the forest, which takes a majority vote of these
classifiers, is itself consistent.

The most recent theoretical study, and the one which achieves the closest match
between theory and practice, is that of \citet{Biau2012}.  The most significant
way in which their model differs from practice is that it requires a second data
set which is not used to fit the leaf predictors in order to make decisions
about variable importance when growing the trees.  One of the innovations of the
model we present in this paper is a way to circumvent this limitation in an
online setting while maintaining consistency.

\section{Online Random Forests with Stream Partitioning}
\label{sec:algorithm-description}

In this section we describe the workings of our online random forest algorithm.
A more precise (pseudo-code) description of the training procedure can be found
in Appendix~\ref{appendix:pseudo-code}. 

\subsection{Forest Construction}

The random forest classifier is constructed by building a collection
of random tree classifiers in parallel.  Each tree is built
independently and in isolation from the other trees in the forest.
Unlike many other random forest algorithms we do not perform
bootstrapping or subsampling at this level; however, the individual
trees each have their own optional mechanism for subsampling the data
they receive.

\subsection{Tree Construction}
Each node of the tree is associated with a rectangular subset of
$\R^D$, and at each step of the construction the collection of cells
associated with the leafs of the tree forms a partition of $\R^D$.
The root of the tree is $\R^D$ itself.  At each step we receive a data
point $(X_t, Y_t)$ from the environment.  Each point is assigned to
one of two possible streams at random with fixed probability.  We
denote stream membership with the variable $I_t \in \{s,e\}$.  How the
tree is updated at each time step depends on which stream the
corresponding data point is assigned to.

We refer to the two streams as the \emph{structure} stream and the
\emph{estimation} stream; points assigned to these streams are
structure and estimation points, respectively.  These names reflect
the different uses of the two streams in the construction of the tree:

\textbf{Structure points} are allowed to influence the structure of
the tree partition, i.e.\ the locations of candidate split points and
the statistics used to choose between candidates, but they are not
permitted to influence the predictions that are made in each leaf of
the tree.

\textbf{Estimation points} are not permitted to influence the shape of
the tree partition, but can be used to estimate class membership
probabilities in whichever leaf they are assigned to.

Only two streams are needed to build a consistent forest, but there is no reason
we cannot have more.  For instance, we explored the use of a third stream for
points that the tree should ignore completely, which gives a form of online
sub-sampling in each tree.  We found empirically that including this third
stream hurts performance of the algorithm, but its presence or absence does not
affect the theoretical properties.

\subsection{Leaf Splitting Mechanism}
\label{subsec:leaf-splitting}
When a leaf is created the number of candidate split dimensions for the new leaf
is set to $\min(1 + \operatorname{Poisson}(\lambda), D)$, and this many distinct
candidate dimensions are selected uniformly at random.  We then collect $m$
candidate splits in each candidate dimension ($m$ is a parameter of the
algorithm) by projecting the first $m$ structure points to arrive in the newly
created leaf onto the candidate dimensions.  We maintain several structural
statistics for each candidate split.  Specifically, for each candidate split we
maintain class histograms for each of the new leafs it would create, using data
from the estimation stream.  We also maintain structural statistics, computed
from data in the structure stream, which can be used to choose between the
candidate splits.  The specific form of the structural statistics does not
affect the consistency of our model, but it is crucial that they depend only on
data in the structure stream.

Finally, we require two additional conditions which control when a leaf at depth
$d$ is split:
\begin{enumerate}
\item Before a candidate split can be chosen, the class histograms in
  each of the leafs it would create must incorporate information from
  at least $\alpha(d)$ estimation points.
\item If any leaf receives more than $\beta(d)$ estimation points, and
  the previous condition is satisfied for \emph{any} candidate split
  in that leaf, then when the next structure point arrives in this leaf it
  must be split regardless of the state of the structural statistics.
\end{enumerate}
The first condition ensures that leafs are not split too often, and the second
condition ensures that no branch of the tree ever stops growing completely.  In
order to ensure consistency we require that $\alpha(d) \to \infty$ monotonically
in $d$ and that $d/\alpha(d) \to 0$.  We also require that $\beta(d) \ge
\alpha(d)$ for convenience.

When a structure point arrives in a leaf, if the first condition is satisfied
for some candidate split then the leaf may optionally be split at the
corresponding point.  The decision of whether to split the leaf or wait to
collect more data is made on the basis of the structural statistics collected
for the candidate splits in that leaf.

\subsection{Structural Statistics}

In each candidate child we maintain an estimate of the posterior
probability of each class, as well as the total number of points we
have seen fall in the candidate child, both counted from the structure
stream.  In order to decide if a leaf should be split, we compute the
information gain for each candidate split which satisfies condition 1
from the previous section,
\begin{align*}
  I(S) = H(A) - \frac{|A'|}{|A|}H(A') -  \frac{|A''|}{|A|}H(A'')
  \enspace.
\end{align*}
Here $S$ is the candidate split, $A$ is the cell belonging to the leaf to be
split, and $A'$ and $A''$ are the two leafs that would be created if $A$ were
split at $S$.  The function $H(A)$ is the discrete entropy, computed over the
labels of the structure points which fall in the cell $A$.

We select the candidate split with the largest information gain for
splitting, provided this split achieves a minimum threshold in
information gain, $\tau$.  The value of $\tau$ is a parameter of our
algorithm.

\subsection{Prediction}
At any time the online forest can be used to make predictions for
unlabelled data points using the model built from the labelled data it
has seen so far.  To make a prediction for a query point $x$ at time
$t$, each tree computes, for each class $k$,
\begin{align*}
  \eta_t^k(x) &= \frac{1}{N^e(A_t(x))}\sum_{\substack{(X_\tau,Y_\tau)\in
      A_t(x)\\I_\tau=e}}\1{Y_\tau = k} 
  \enspace,
\end{align*}
where $A_t(x)$ denotes the leaf of the tree containing $x$ at time
$t$, and $N^e(A_t(x))$ is the number of estimation points which have
been counted in $A_t(x)$ during its lifetime.  Similarly, the sum is
over the labels of these points.  The tree prediction is then the
class which maximizes this value:
\begin{align*}
  g_t(x) = \arg\max_k\{\eta_t^k(x)\}
  \enspace.
\end{align*}
The forest predicts the class which receives the most votes from
the individual trees.

Note that this requires that we maintain class histograms 
from both the structure and estimation streams separately for each
candidate child in the fringe of the tree.  The counts from the
structure stream are used to select between candidate split points,
and the counts from the estimation stream are used to initialize the
parameters in the newly created leafs after a split is made.

\subsection{Memory Management}
\label{ssec:memory-management}


The typical approach to building trees online, which is employed in
\citet{domingos2000mining} and \citet{saffari2009line}, is to maintain a fringe
of candidate children in each leaf of the tree.  The algorithm collects
statistics in each of these candidate children until some (algorithm dependent)
criterion is met, at which point a pair of candidate children is selected to
replace their parent.  The selected children become leafs in the new tree,
acquiring their own candidate children, and the process repeats.  Our algorithm
also uses this approach.

The difficulty here is that the trees must be grown breadth first, and
maintaining the fringe of potential children is very memory intensive when the
trees are large.  Our algorithm also suffers from this deficiency, as
maintaining the fringe requires $O(cmd)$ statistics in each leaf, where $d$ is
the number of candidate split dimensions, $m$ is the number of candidate split
points (i.e.\ $md$ pairs of candidate children per leaf) and $c$ is the number
of classes in the problem.  These statistics can be quite large and for deep
trees the memory cost becomes prohibitive.


In practice the memory problem is resolved either by growing small trees, as in
\citet{saffari2009line}, or by bounding the number of nodes in the fringe of the
tree, as in \citet{domingos2000mining}.  Other models of streaming random
forests, such as those discussed in \citet{Abdulsalam2008}, build trees in
sequence instead of in parallel, which reduces the total memory usage.


Our algorithm makes use of a bounded fringe and adopts the technique of
\citet{domingos2000mining} to control the policy for adding and removing leafs
from the fringe.

In each tree we partition the leafs into two sets: we have a set of
\emph{active} leafs, for which we collect split statistics as described in
earlier sections, and a set of \emph{inactive} leafs for which we store only two
numbers.  We call the set of active leafs the \emph{fringe} of the tree, and
describe a policy for controlling how inactive leafs are added to the fringe.

In each inactive leaf $A_t$ we store the following two quantities
\begin{itemize}
\item $\hat{p}(A_t)$ which is an estimate of $\mu(A_t) = \P{X \in A_t}$, and
\item $\hat{e}(A_t)$ which is an estimate of $e(A) = \P{g_t(X) \neq Y \,|\, X
    \in A_t}$.
\end{itemize}
Both of these are estimated based on the estimation points which arrive in $A_t$
during its lifetime.  From these two numbers we form the statistic $\hat{s}(A_t)
= \hat{p}(A_t)\hat{e}(A_t)$ (with corresponding true value $s(A_t) =
p(A_t)e(A_t)$) which is an upper bound on the improvement in error rate that can
be obtained by splitting $A_t$.

Membership in the fringe is controlled by $\hat{s}(A_t)$.  When a leaf is split
it relinquishes its place in the fringe and the inactive leaf with the largest
value of $\hat{s}(A_t)$ is chosen to take its place.  The newly created leafs
from the split are initially inactive and must compete with the other inactive
leafs for entry into the fringe.

Unlike \citet{domingos2000mining}, who use this technique only as a heuristic
for managing memory use, we incorporate the memory management directly into our
analysis.  The analysis in Appendix~\ref{appendix:consistency} shows that our
algorithm, including a limited size fringe, is consistent.


\section{Theory}
\label{sec:theory}

In this section we state our main theoretical results and give an
outline of the strategy for establishing consistency of our online
random forest algorithm.  In the interest of space and clarity we do
not include proofs in this section.  Unless otherwise noted, the
proofs of all claims appear in Appendix~\ref{appendix:consistency}.

We denote the tree partition created by our online random forest
algorithm from $t$ data points as $g_t$.  As $t$ varies we obtain a
sequence of classifiers, and we are interested in showing that the
sequence $\{g_t\}$ is consistent, or more precisely that the
probability of error of $g_t$ converges in probability to the Bayes
risk, i.e.\
\begin{align*}
  L(g_t) = \P{g_t(X,Z) \neq Y \,|\, D_t} \to L^*
  \enspace,
\end{align*}
as $t\to\infty$.  Here $(X,Y)$ is a random test point and $Z$ denotes the
randomness in the tree construction algorithm.  $D_t$ is the training set (of
size $t$) and the probability in the convergence is over the random selection of
$D_t$.  The Bayes risk is the probability of error of the Bayes classifier,
which is the classifier that makes predictions by choosing the class with the
highest posterior probability,
\begin{align*}
  g(x) = \arg\max_k \P{Y=k\,|\,X=x}
  \enspace,
\end{align*}
(where ties are broken in favour of the smaller index).  The Bayes
risk $L(g)=L^*$ is the minimum achievable risk of any classifier for
the distribution of $(X,Y)$.  In order to ease notation, we drop the
explicit dependence on $D_t$ in the remainder of this paper.  More
information about this setting can be found in \citet{devroye96}.

Our main result is the following theorem:

\begin{theorem}
  Suppose the distribution of $X$ has a density with respect to the
  Lebesgue measure and that this density is bounded from above and
  below.  Then the online random forest classifier described in this
  paper is consistent.
  \label{thm:consistent-forest}
\end{theorem}
The first step in proving Theorem~\ref{thm:consistent-forest} is to
show that the consistency of a voting classifier, such as a random
forest, follows from the consistency of the base classifiers.  We
prove the following proposition, which is a straightforward
generalization of a proposition from~\citet{biau08}, who prove the
same result for two class ensembles.
\begin{proposition}
  Assume that the sequence $\{g_t\}$ of randomized classifiers is consistent for
  a certain distribution of $(X, Y)$.  Then the voting classifier, $g_t^{(M)}$
  obtained by taking the majority vote over $M$ (not necessarily independent)
  copies of $g_t$ is also consistent.
  \label{prop:multi-biau}
\end{proposition}
With Proposition~\ref{prop:multi-biau} established, the remainder of
the effort goes into proving the consistency of our tree construction.

The first step is to separate the stream splitting randomness from the
remaining randomness in the tree construction.  We show that if a
classifier is conditionally consistent based on the outcome of some
random variable, and the sampling process for this random variable
generates acceptable values with probability 1, then the resulting
classifier is unconditionally consistent.

\begin{proposition}
  Suppose $\{g_t\}$ is a sequence of classifiers whose probability of
  error converges conditionally in probability to the Bayes risk $L^*$
  for a specified distribution on $(X, Y)$, i.e.\
  \begin{align*}
    \P{g_t(X,Z,I)\neq Y\,|\,I} \to L^*
  \end{align*}
  for all $I \in \mathcal{I}$ and that $\nu$ is a distribution on $I$.
  If $\nu(\mathcal{I}) = 1$ then the probability of error converges
  unconditionally in probability, i.e.\
  \begin{align*}
    \P{g_t(X,Z,I)\neq Y} \to L^*
  \end{align*}
  In particular, $\{g_t\}$ is consistent for the specified distribution.
  \label{prop:condition-on-full-measure}
\end{proposition}
Proposition~\ref{prop:condition-on-full-measure} allows us to
condition on the random variables $\{I_t\}_{t=1}^\infty$ which
partition the data stream into structure and estimation points in each
tree.  Provided that the random partitioning process produces
acceptable sequences with probability 1, it is sufficient to show that
the random tree classifier is consistent conditioned on such a
sequence.  In particular, in the remainder of the argument we assume
that $\{I_t\}_{t=1}^\infty$ is a fixed, deterministic sequence which
assigns infinitely many points to each of the structure and estimation
streams.  We refer to such a sequence as a \emph{partitioning
  sequence}.

\begin{figure}[tbh]
  \centering
  \begin{tikzpicture}
    [every node/.style={draw=black, circle}]
    \node (S) at (-2, 0) {$S$};
    \node (I) at (0, 0) [fill=black!30] {$I$};
    \node (E) at (2, 0) {$E$};
    \draw (S) -- (I) -- (E);
  \end{tikzpicture}
  \caption{The dependency structure of our algorithm.  $S$ represents
    the randomness in the structure of the tree partition, $E$
    represents the randomness in the leaf estimators and $I$
    represents the randomness in the partitioning of the data stream.
    $E$ and $S$ are independent conditioned on $I$.}
  \label{fig:dependency-structure}
\end{figure}
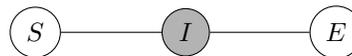

The reason this is useful is that conditioning on a partitioning
sequence breaks the dependence between the structure of the tree
partition and the estimators in the leafs.  This is a powerful tool
because it gives us access to a class of consistency theorems which
rely on this type of independence.  However, before we are able to
apply these theorems we must further reduce our problem to proving the
consistency of estimators of the posterior distribution of each class.

\begin{proposition}
  Suppose we have regression estimates, $\eta_t^k(x)$, for each class
  posterior $\eta^k(x) = \P{Y=k \,|\, X=x}$, and that these estimates are each
  consistent.  The classifier
  \begin{align*}
    g_t(x) = \arg\max_k\{\eta_t^k(x)\}
  \end{align*}
  (where ties are broken in favour of the smaller index) is consistent
  for the corresponding multiclass classification problem.
  \label{prop:multiclass}
\end{proposition}

Proposition~\ref{prop:multiclass} allows us to reduce the consistency
of the multiclass classifier to the problem of proving the consistency
of several two class posterior estimates.  Given a set of classes
$\{1,\ldots,c\}$ we can re-assign the labels using the map $(X, Y)
\mapsto (X, \1{Y=k})$ for any $k\in\{1,\ldots,c\}$ in order to get a
two class problem where $\P{Y=1 \,|\, X=x}$ in this new problem is
equal to $\eta^k(x)$ in the original multiclass problem.  Thus to
prove consistency of the multiclass classifier it is enough to show
that each of these two class posteriors is consistent.  To this end we
make use of the following theorem from~\citet{devroye96}.

\begin{theorem}
  Consider a partitioning classification rule which builds a
  prediction $\eta_t(x)$ of $\eta(x) = \P{Y=1\,|\,X=x}$ by averaging
  the labels in each cell of the partition.  If the labels of the
  voting points do not influence the structure of the partition then
  \begin{align*}
    \E{|\eta_t(x) - \eta(x)|} \to 0
  \end{align*}
  provided that
  \begin{enumerate}
  \item $\diam(A_t(X)) \to 0$ in probability,
  \item $N^e(A_t(X)) \to \infty$ in probability.
  \end{enumerate}
  \label{thm:devroye61}
\end{theorem}
\begin{proof}
  See Theorem 6.1 in~\citet{devroye96}.
\end{proof}

Here $A_t(X)$ refers to the cell of the tree partition containing a random test
point $X$, and $\diam(A)$ indicates the diameter of set $A$, which is defined as
the maximum distance between any two points falling in $A$,
\begin{align*}
  \diam(A) = \sup_{x,y\in A} ||x - y||
  \enspace.
\end{align*}
The quantity $N^e(A_t(X))$ is the number of points contributing to the
estimation of the posterior at $X$.

This theorem places two requirements on the cells of the partition.  The first
condition ensures that the cells are sufficiently small that small details of
the posterior distribution can be represented.  The second condition requires
that the cells be large enough that we are able to obtain high quality estimates
of the posterior probability in each cell.

The leaf splitting mechanism described in
Section~\ref{subsec:leaf-splitting} ensures that the second condition
of Theorem~\ref{thm:devroye61} is satisfied.  However, showing that
our algorithm satisfies the first condition requires significantly
more work.  The chief difficulty lies in showing that every leaf of
the tree will be split infinitely often in probability.  Once
this claim is established a relatively straightforward calculation
shows that the expected size of each dimension of a leaf is reduced
each time it is split.

So far we have described the approach to proving consistency of our algorithm
with an unbounded fringe.  If the tree is small (i.e.\ never has more leafs than
the maximum fringe size) then the analysis is unchanged.  However, since our
trees are required to grow to unbounded size this is not possible.

In order to apply Theorem~\ref{thm:devroye61} in the case of an unbounded fringe
we have shown that every leaf will be split in finite time with arbitrarily high
probability.  To extend consistency to this setting we need only show that the
probability of an inactive leaf not being activated goes to zero as
$t\to\infty$.  This is sufficient, since once a leaf is activated it remains in
the fringe until it is split and the argument from the unbounded fringe setting
applies.

In order to show that any leaf will be eventually added to the fringe, we
consider an arbitrary leaf $A$ and show that we can make the probability that
$\hat{s}(A)$ is not the largest $\hat{s}(A)$ value among inactive leafs
arbitrarily small by making $t$ sufficiently large.

These details are somewhat lengthy, so we refer the interested reader to
Appendix~\ref{appendix:consistency} a full presentation, including proofs of the
propositions stated in this section.


\section{Experiments}
\label{sec:experiments}

In this section we demonstrate some empirical results in order to illustrate the
properties of our algorithm.
Code to reproduce all of the experiements in this section is available
online\footnote{\url{https://github.com/david-matheson/rftk-colrf-icml2013}}.

\begin{figure}[tbh]
  \centering
  \includegraphics[width=0.9\linewidth]{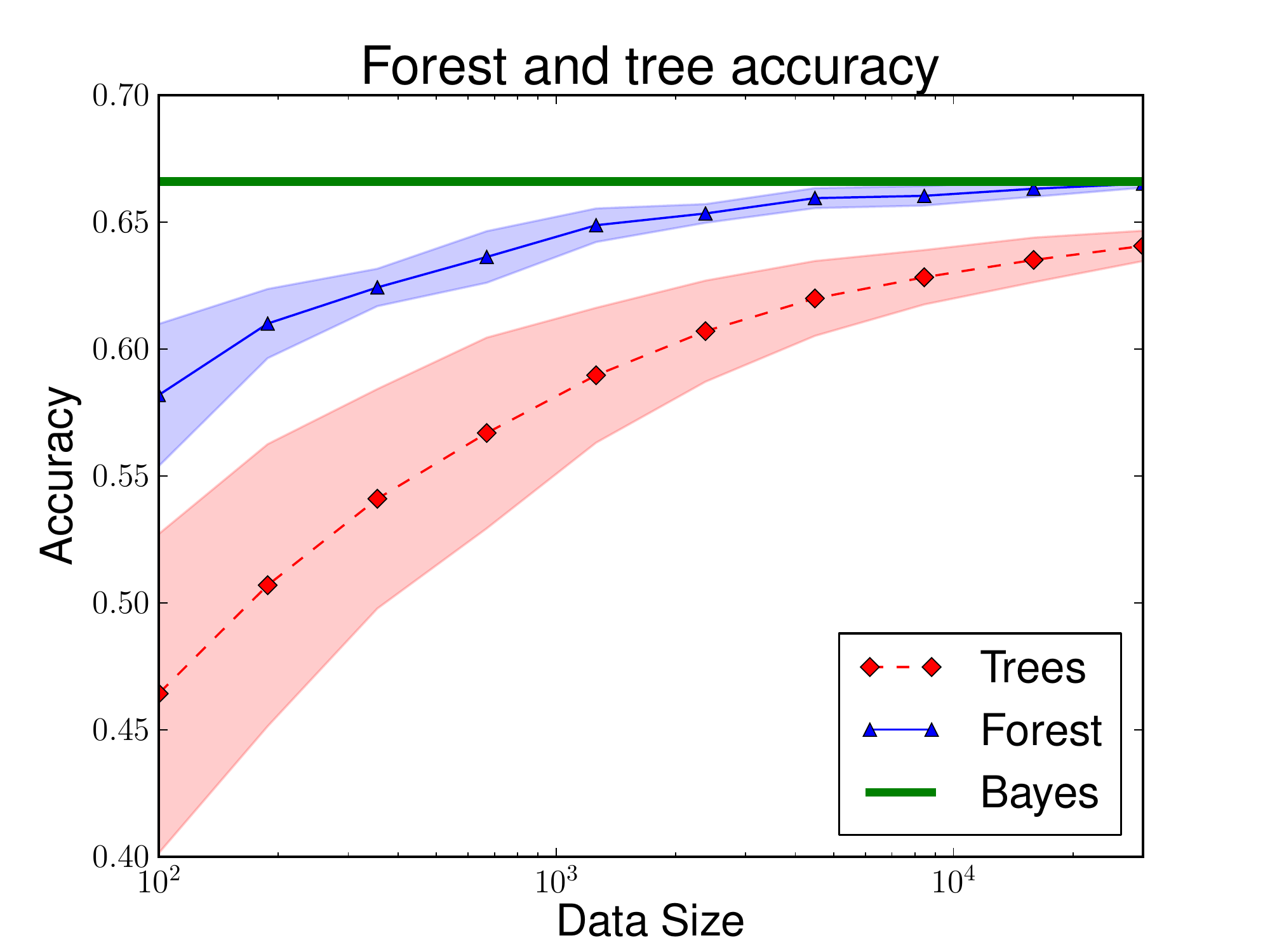}
  \caption{Prediction accuracy of the forest and the trees it averages on a 2D
    mixture of Gaussians.  The horizontal line shows the accuracy of the Bayes
    classifier on this problem.  We see that the accuracy of the forest
    consistently dominates the expected accuracy of the trees.  Shaded regions
    show one standard deviation computed over 10 runs.}
  \label{fig:forest-and-trees}
\end{figure}

\subsection{Advantage of a Forest}
Our first experiment demonstrates that although the individual trees are
consistent classifiers, empirically the performance of the forest is
significantly better than each of the trees for problems with finite data.  We
demonstrate this on a synthetic five class mixture of Gaussians problem with
significant class overlap and variation in prior weights.  For this experiment
we used 100 trees and set $\lambda=1$, $m=10$, $\tau=0.001$, $\alpha(d) =
1.1^d$, $\beta(d)=1000\alpha(d)$.

From Figure~\ref{fig:forest-and-trees} it is clear that the forest
converges much more quickly than the individual trees.  Result
profiles of this kind are common in the boosting and random forests
literature; however, in practice one often uses inconsistent base
classifiers in the ensemble (e.g.\ boosting with decision stumps or
random forests where the trees are grown to full size).  This
experiment demonstrates that although our base classifiers provably
converge, empirically there is still a benefit from averaging in
finite time.

\subsection{Comparison to Offline}

In our second experiment, we demonstrate that our online algorithm is
able to achieve similar performance to an offline implementation of
random forests and also compare to an existing online random forests
algorithm on a small non-synthetic problem.

In particular, we demonstrate this on the USPS data set from the LibSVM
repository \cite{CC01a}.  We have chosen the USPS data for this experiment
because it allows us to compare our results directly to those of
\citet{saffari2009line}, whose algorithm is very similar to our own.  For both
algorithms we use a forest of 100 trees.  For our model we set $\lambda=10$,
$m=10$, $\tau=0.1$, $\alpha(d)=10(1.00001^d)$ and $\beta(d)=10^4\alpha(d)$.  For
the model of \citet{saffari2009line} we set the number of features and
thresholds to sample at 10, the minimum information gain to 0.1 and the minimum
number of samples to split at leaf at 50.  We show results from both online
algorithms with 15 passes through the data.

\begin{figure}[tb]
  \centering
  \includegraphics[width=0.9\linewidth]{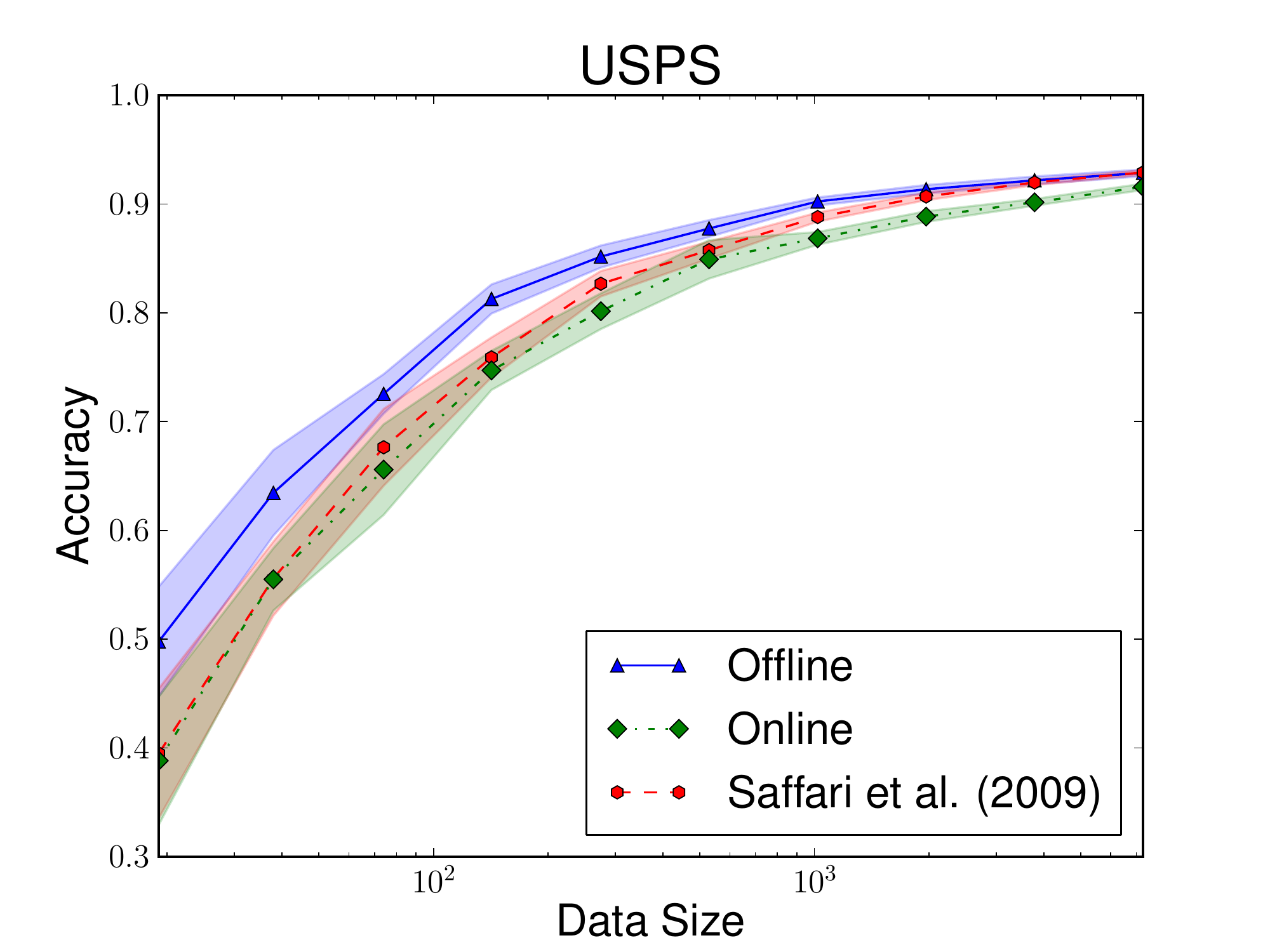}
  \caption{Comparison between offline random forests and our online algorithm on
    the USPS data set.  The online forest uses 10 passes through the data set.
    The third line is our implementation of the algorithm from
    \citet{saffari2009line}.  Shaded regions show one standard deviation
    computed over 10 runs.}
  \label{fig:usps}
\end{figure}

Figure~\ref{fig:usps} shows that we are able to achieve performance very similar
to the offline random forest on the full data.  The performance we achieve is
similar to the performance reported by \citet{saffari2009line} on this data set.

\subsection{Microsoft Kinect}

For our final experiment we evaluate our online random forest algorithm on the
challenging computer vision problem of predicting human body part labels from a
depth image. Our procedure closely follows the work of \citet{Shotton2011} which
is used in the commercially successful Kinect system. Applying the same approach
as \citet{Shotton2011}, our online classifier predicts the body part label of a
single pixel $P$ in a depth image.  To predict all the labels of a depth image,
the classifier is applied to every pixel in parrallel.

\begin{figure}[tb]
\centering
  \vskip4ex
  \raisebox{0.4cm}{\includegraphics[height=0.2\textheight]{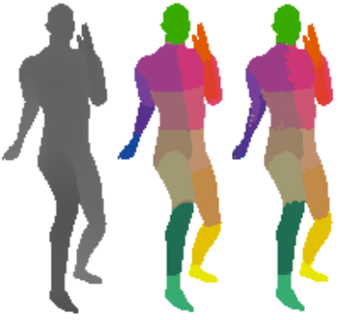}}
  ~~~~~~~~~~
  \raisebox{0.4cm}{\includegraphics[height=0.2\textheight]{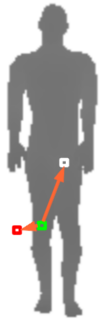}}
  \caption{\textbf{Left:} Depth, ground truth body parts and predicted body
    parts. \textbf{Right:} A candidate feature specified by two offsets.}
  \label{fig:kinect-example}
\end{figure}

For our dataset, we generate pairs of 640x480 resolution depth and body part
images by rendering random poses from the CMU mocap dataset. The 19 body parts
and one background class are represented by 20 unique color identifiers in the
body part image. Figure \ref{fig:kinect-example}~(left) visualizes the raw depth
image, ground truth body part labels and body parts predicted by our classifier
for one pose. During training, we sample 50 pixels without replacement for each
body part class from each pose; thus, producing 1000 data points for each depth
image. During testing we evaluate the prediction accuracy of all non background
pixels as this provides a more informative accuracy metric since most of the
pixels are background and are relatively easy to predict. For this experiment we
use a stream of 2000 poses for training and 500 poses for testing.

Each node of each decision tree computes the depth difference between two pixels
described by two offsets from $P$ (the pixel being classified). At training
time, candidate pairs of offsets are sampled from a 2-dimensional Gaussian
distributions with variance 75.0. The offsets are scaled by the depth of the
pixel $P$ to produce depth invariant features. Figure
\ref{fig:kinect-example}~(right) shows a candidate feature for the indicated
pixel. The resulting feature value is the depth difference between the pixel in
the red box and the pixel in the white box.

In this experiment we construct a forest of 25 trees with 2000 candidate offsets
($\lambda$), 10 candidate splits ($m$) and a minimum information gain of 0.01
($\tau$). For \citet{saffari2009line} we set the number of sample points
required to split to 25 and for our own algorithm we set
$\alpha(d)=25\cdot(1.01^d)$ and $\beta(d)=4\cdot\alpha(d)$. With this parameter
setting each active leaf stores $20\cdot10\cdot2000\cdot2=400,000$ statistics
which requires 1.6MB of memory. By limiting the fringe to 1000 active leaves our
algorithm requires 1.6GB of memory for leaf statistics. To limit the maximum
memory used by \citet{saffari2009line} we set the maximum depth to 8 which uses
up to $25\cdot2^8=6400$ active leaves which requires up to 10GB of memory for
leaf statistics.

\begin{figure}[tb]
  \centering
  \includegraphics[width=0.9\linewidth]{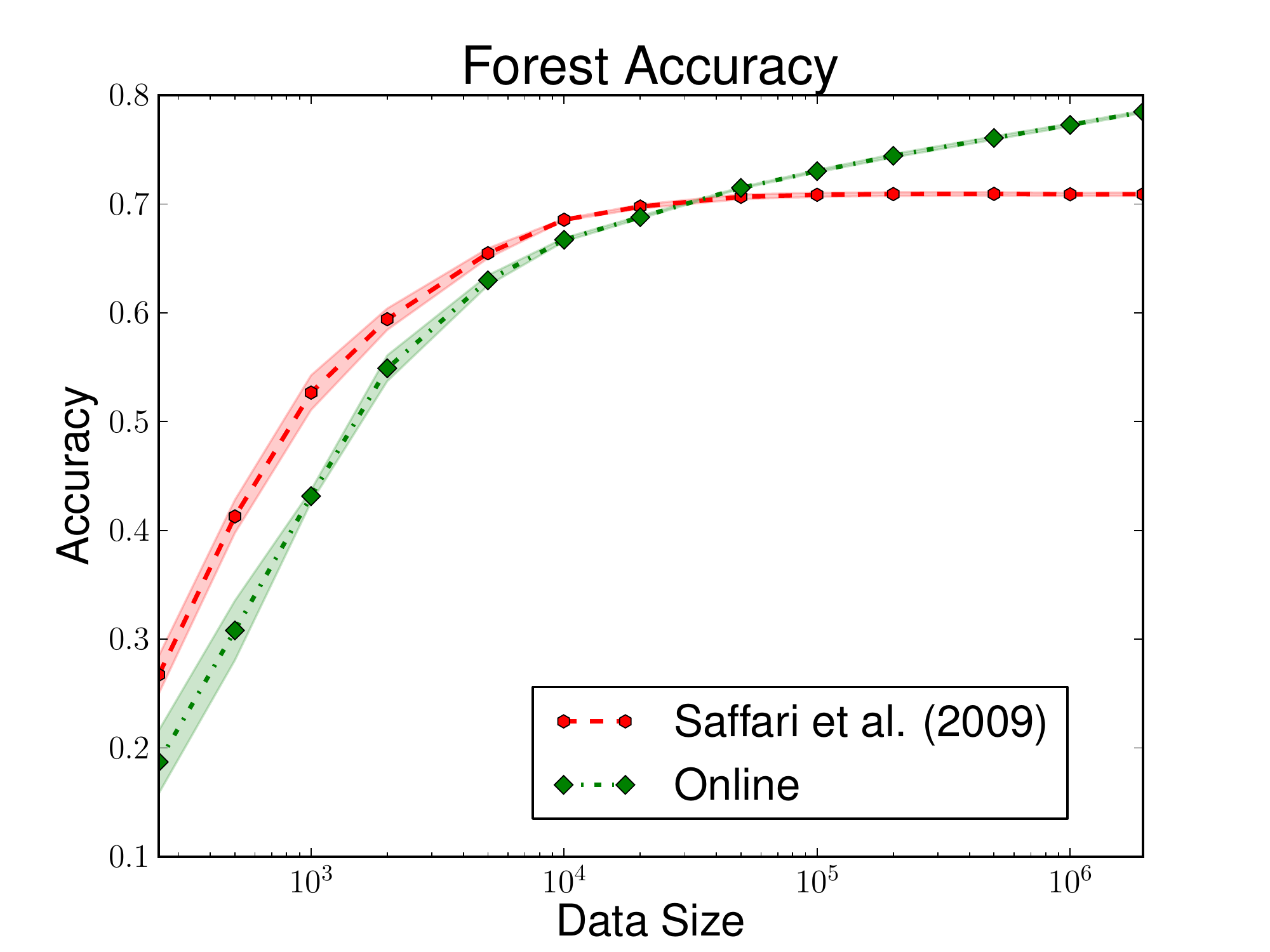}
  \caption{Comparison of our online algorithm with \citet{saffari2009line} on
    the kinect application. Error regions show one standard deviation computed
    over 5 runs.}
  \label{fig:kinect}
\end{figure}

Figure~\ref{fig:kinect} shows that our algorithm achieves significantly better
accuracy while requiring less memory.

\section{Discussion and Future Work}
\label{sec:conclusion}

In this paper we described an algorithm for building online random
forests and showed that our algorithm is consistent.  To the best of
our knowledge this is the first consistency result for online random
forests.


Growing trees online in the obvious way requires large amounts of memory, since
the trees must be grown breadth first and each leaf must store are large number
of statistics in each of its potential children.  We incorporated a memory
management technique from \citet{domingos2000mining} in order to limit the
number of leafs in the fringe of the tree.  This refinement is important, since
it enables our algorithm to grow large trees.  The analysis shows that our
algorithm is still consistent with this refinement.

The analysis we presented in this paper shows that our algorithm is consistent,
but does not give rates of convergence to the Bayes risk.  Analyzing the
convergence rate of our algorithm is clear direction for future work, but the
way to proceed does not appear to be straightforward.

Finally, our current algorithm is restricted to axis aligned splits.
Many implementations of random forests use more elaborate split
shapes, such as random linear or quadratic combinations of features.
These strategies can be highly effective in practice, especially in
sparse or high dimensional settings.  Understanding how to maintain
consistency in these settings is another potentially interesting
direction of inquiry.


\vspace{-0.1cm}
\section*{Acknowledgements}
\vspace{-0.1cm}

Some of the data used in this paper was obtained from \url{mocap.cs.cmu.edu}
(funded by NSF EIA-0196217).


{

\setlength\bibsep{0.3cm} 

\small
\bibliography{bayesopt,randomforests}

\begin{thebibliography}{8}
\providecommand{\natexlab}[1]{#1}
\providecommand{\url}[1]{\texttt{#1}}
\expandafter\ifx\csname urlstyle\endcsname\relax
  \providecommand{\doi}[1]{doi: #1}\else
  \providecommand{\doi}{doi: \begingroup \urlstyle{rm}\Url}\fi

\bibitem[Author(2011)]{anonymous}
Author, N.~N.
\newblock Suppressed for anonymity, 2011.

\bibitem[Duda et~al.(2000)Duda, Hart, and Stork]{DudaHart2nd}
Duda, R.~O., Hart, P.~E., and Stork, D.~G.
\newblock \emph{Pattern Classification}.
\newblock John Wiley and Sons, 2nd edition, 2000.

\bibitem[Kearns(1989)]{kearns89}
Kearns, M.~J.
\newblock \emph{Computational Complexity of Machine Learning}.
\newblock PhD thesis, Department of Computer Science, Harvard University, 1989.

\bibitem[Langley(2000)]{langley00}
Langley, P.
\newblock Crafting papers on machine learning.
\newblock In Langley, Pat (ed.), \emph{Proceedings of the 17th International
  Conference on Machine Learning (ICML 2000)}, pp.\  1207--1216, Stanford, CA,
  2000. Morgan Kaufmann.

\bibitem[Michalski et~al.(1983)Michalski, Carbonell, and
  Mitchell]{MachineLearningI}
Michalski, R.~S., Carbonell, J.~G., and Mitchell, T.~M. (eds.).
\newblock \emph{Machine Learning: An Artificial Intelligence Approach, Vol. I}.
\newblock Tioga, Palo Alto, CA, 1983.

\bibitem[Mitchell(1980)]{mitchell80}
Mitchell, T.~M.
\newblock The need for biases in learning generalizations.
\newblock Technical report, Computer Science Department, Rutgers University,
  New Brunswick, MA, 1980.

\bibitem[Newell \& Rosenbloom(1981)Newell and Rosenbloom]{Newell81}
Newell, A. and Rosenbloom, P.~S.
\newblock Mechanisms of skill acquisition and the law of practice.
\newblock In Anderson, J.~R. (ed.), \emph{Cognitive Skills and Their
  Acquisition}, chapter~1, pp.\  1--51. Lawrence Erlbaum Associates, Inc.,
  Hillsdale, NJ, 1981.

\bibitem[Samuel(1959)]{Samuel59}
Samuel, A.~L.
\newblock Some studies in machine learning using the game of checkers.
\newblock \emph{IBM Journal of Research and Development}, 3\penalty0
  (3):\penalty0 211--229, 1959.

\end{thebibliography}


\begin{thebibliography}{26}
\providecommand{\natexlab}[1]{#1}
\providecommand{\url}[1]{\texttt{#1}}
\expandafter\ifx\csname urlstyle\endcsname\relax
  \providecommand{\doi}[1]{doi: #1}\else
  \providecommand{\doi}{doi: \begingroup \urlstyle{rm}\Url}\fi

\bibitem[Abdulsalam(2008)]{Abdulsalam2008}
H.~Abdulsalam.
\newblock \emph{{Streaming Random Forests}}.
\newblock PhD thesis, Queens University, 2008.

\bibitem[Biau(2012)]{Biau2012}
G.~Biau.
\newblock {Analysis of a Random Forests model}.
\newblock \emph{JMLR}, 13\penalty0 (April):\penalty0 1063--1095, 2012.

\bibitem[Biau et~al.(2008)Biau, Devroye, and Lugosi]{biau08}
G.~Biau, L.~Devroye, and G.~Lugosi.
\newblock Consistency of random forests and other averaging classifiers.
\newblock \emph{JMLR}, 9:\penalty0 2015--2033, 2008.

\bibitem[Bifet et~al.(2010)Bifet, Holmes, and Pfahringer]{Bifet2010}
A.~Bifet, G.~Holmes, and B.~Pfahringer.
\newblock {MOA: Massive Online Analysis, a framework for stream classification
  and clustering.}
\newblock In \emph{Workshop on Applications of Pattern Analysis}, pp.\  3--16,
  2010.

\bibitem[Bifet et~al.(2012)Bifet, Frank, Holmes, and Pfahringer]{Bifet2012}
A.~Bifet, E.~Frank, G.~Holmes, and B.~Pfahringer.
\newblock {Ensembles of Restricted Hoeffding Trees}.
\newblock \emph{ACM Transactions on Intelligent Systems and Technology},
  3\penalty0 (2):\penalty0 1--20, February 2012.

\bibitem[Bifet et~al.(2009)Bifet, Holmes, and Pfahringer]{Bifet2009}
A.~Bifet, G.~Holmes, and B.~Pfahringer.
\newblock {New ensemble methods for evolving data streams}.
\newblock In \emph{ACM SIGKDD Intl. Conference on Knowledge Discovery and Data
  Mining}, 2009.

\bibitem[Bosch et~al.(2007)Bosch, Zisserman, and Munoz]{Bosch}
A.~Bosch, A.~Zisserman, and X.~Munoz.
\newblock Image classification using random forests and ferns.
\newblock In \emph{International Conference on Computer Vision}, pp.\  1--8,
  2007.

\bibitem[Breiman(2001)]{Breiman2001:random_forests}
L.~Breiman.
\newblock Random forests.
\newblock \emph{Machine Learning}, 45\penalty0 (1):\penalty0 5--32, 2001.

\bibitem[Breiman(2004)]{Breiman2004}
L.~Breiman.
\newblock {Consistency for a Simple Model of Random Forests}.
\newblock Technical report, University of California at Berkeley, 2004.

\bibitem[Breiman et~al.(1984)Breiman, Friedman, Stone, and Olshen]{breiman84}
L.~Breiman, J.~Friedman, C.~Stone, and R.~Olshen.
\newblock \emph{Classification and Regression Trees}.
\newblock CRC Press LLC, Boca Raton, Florida, 1984.

\bibitem[Chang \& Lin(2011)Chang and Lin]{CC01a}
C.~Chang and C.~Lin.
\newblock {LIBSVM}: A library for support vector machines.
\newblock \emph{ACM Transactions on Intelligent Systems and Technology},
  2:\penalty0 27:1--27:27, 2011.

\bibitem[Criminisi et~al.(2011)Criminisi, Shotton, and
  Konukoglu]{criminisi2011decision}
A.~Criminisi, J.~Shotton, and E.~Konukoglu.
\newblock Decision forests: A unified framework for classification, regression,
  density estimation, manifold learning and semi-supervised learning.
\newblock \emph{Foundations and Trends in Computer Graphics and Vision},
  7\penalty0 (2-3):\penalty0 81--227, 2011.

\bibitem[Cutler et~al.(2007)Cutler, Edwards, and Beard]{Cutler2007}
D.~Cutler, T.~Edwards, and K.~Beard.
\newblock {Random forests for classification in ecology}.
\newblock \emph{Ecology}, 88\penalty0 (11):\penalty0 2783--92, November 2007.

\bibitem[Devroye et~al.(1996)Devroye, Gy{\"o}rfi, and Lugosi]{devroye96}
L.~Devroye, L.~Gy{\"o}rfi, and G.~Lugosi.
\newblock \emph{A Probabilistic Theory of Pattern Recognition}.
\newblock Springer-Verlag, New York, USA, 1996.

\bibitem[Domingos \& Hulten(2000)Domingos and Hulten]{domingos2000mining}
P.~Domingos and G.~Hulten.
\newblock Mining high-speed data streams.
\newblock In \emph{International Conference on Knowledge Discovery and Data
  Mining}, pp.\  71--80. ACM, 2000.

\bibitem[Gama et~al.(2005)Gama, Medas, and
  Rodrigues]{Gama:2005:LDT:1066677.1066809}
J.~Gama, P.~Medas, and P.~Rodrigues.
\newblock Learning decision trees from dynamic data streams.
\newblock In \emph{ACM symposium on Applied computing}, SAC '05, pp.\
  573--577, New York, NY, USA, 2005. ACM.

\bibitem[Genuer(2010)]{genuer2010risk}
R.~Genuer.
\newblock Risk bounds for purely uniformly random forests.
\newblock Technical report, Institut National de Recherche en Informatique et
  en Automatique, 2010.

\bibitem[Genuer(2012)]{genuer2012variance}
R.~Genuer.
\newblock Variance reduction in purely random forests.
\newblock \emph{Journal of Nonparametric Statistics}, 24\penalty0 (3):\penalty0
  543--562, 2012.

\bibitem[Ishwaran \& Kogalur(2010)Ishwaran and Kogalur]{Ishwaran2010}
H.~Ishwaran and U.~Kogalur.
\newblock {Consistency of random survival forests}.
\newblock \emph{Statistics and Probability Letters}, 80:\penalty0 1056--1064,
  2010.

\bibitem[Lin \& Jeon(2002)Lin and Jeon]{Lin2002}
Y.~Lin and Y.~Jeon.
\newblock {Random forests and adaptive nearest neighbors}.
\newblock Technical Report 1055, University of Wisconsin, 2002.

\bibitem[Meinshausen(2006)]{Meinshausen2006}
N.~Meinshausen.
\newblock {Quantile regression forests}.
\newblock \emph{JMLR}, 7:\penalty0 983--999, 2006.

\bibitem[Oza \& Russel(2001)Oza and Russel]{Oza2001}
N.~Oza and S.~Russel.
\newblock {Online Bagging and Boosting}.
\newblock In \emph{Artificial Intelligence and Statistics}, volume~3, 2001.

\bibitem[Prasad et~al.(2006)Prasad, Iverson, and Liaw]{Prasad2006}
A.~Prasad, L.~Iverson, and A.~Liaw.
\newblock {Newer Classification and Regression Tree Techniques: Bagging and
  Random Forests for Ecological Prediction}.
\newblock \emph{Ecosystems}, 9\penalty0 (2):\penalty0 181--199, March 2006.
\newblock ISSN 1432-9840.

\bibitem[Saffari et~al.(2009)Saffari, Leistner, Santner, Godec, and
  Bischof]{saffari2009line}
A.~Saffari, C.~Leistner, J.~Santner, M.~Godec, and H.~Bischof.
\newblock On-line random forests.
\newblock In \emph{International Conference on Computer Vision Workshops (ICCV
  Workshops)}, pp.\  1393--1400. IEEE, 2009.

\bibitem[Shotton et~al.(2011)Shotton, Fitzgibbon, Cook, Sharp, Finocchio,
  Moore, Kipman, and Blake]{Shotton2011}
J.~Shotton, A.~Fitzgibbon, M.~Cook, T.~Sharp, M.~Finocchio, R.~Moore,
  A.~Kipman, and A.~Blake.
\newblock {Real-time human pose recognition in parts from single depth images}.
\newblock \emph{CVPR}, pp.\  1297--1304, 2011.

\bibitem[Svetnik et~al.(2003)Svetnik, Liaw, Tong, Culberson, Sheridan, and
  Feuston]{Svetnik2003}
V.~Svetnik, A.~Liaw, C.~Tong, J.~Culberson, R.~Sheridan, and B.~Feuston.
\newblock {Random forest: a classification and regression tool for compound
  classification and QSAR modeling.}
\newblock \emph{Journal of Chemical Information and Computer Sciences},
  43\penalty0 (6):\penalty0 1947--58, 2003.

\end{thebibliography}
\bibliographystyle{icml2013}
}

\appendix

\clearpage
\onecolumn
\section{Algorithm pseudo-code}
\label{appendix:pseudo-code}

We present pseudo-code for the basic algorithm only, without the bounded fringe
technique described in Section~\ref{ssec:memory-management}.  The addition of a
bounded fringe is straightforward, but complicates the presentation
significantly.

\vspace{-0.4cm}
\begin{description}
\item[Candidate split dimension] A dimension along which a split may be made.
  \vspace{-0.1cm}
\item[Candidate split point] One of the first $m$ structure points to arrive in
  a leaf.
  \vspace{-0.1cm}
\item[Candidate split] A combination of a candidate split dimension and a
  position along that dimension to split.  These are formed by projecting each
  candidate split point into each candidate split dimension.
  \vspace{-0.1cm}
\item[Candidate children] Each candidate split in a leaf induces two candidate
  children for that leaf.  These are also referred to as the left and right
  child of that split.
  \vspace{-0.1cm}
\item[$N^e(A)$] is a count of estimation points in the cell $A$, and
  $Y^e(A)$ is the histogram of labels of these points in $A$.  $N^s(A)$ and
  $Y^s(A)$ are the corresponding values derived from structure points.
\end{description}
\vspace{-0.5cm}

\begin{algorithm}
\caption{BuildTree}
\begin{algorithmic}
  \REQUIRE Initially the tree has exactly one leaf (TreeRoot) which covers the
  whole space
  \REQUIRE The dimensionality of the input, $D$.  Parameters $\lambda$, $m$ and $\tau$.
  \STATE SelectCandidateSplitDimensions(TreeRoot,
  $\min(1+\operatorname{Poisson}(\lambda), D)$)
  \FOR{$t = 1 \ldots$} 
    \STATE Receive $(X_t, Y_t, I_t)$ from the environment
    \STATE $A_t \leftarrow$ leaf containing $X_t$
    \IF {$I_t = \text{estimation}$}
      \STATE UpdateEstimationStatistics($A_t$, $(X_t, Y_t)$)
      \FORALL {$S \in \operatorname{CandidateSplits}(A_t)$}
        \FORALL {$A \in \operatorname{CandidateChildren}(S)$}
          \IF {$X_t \in A$}
            \STATE UpdateEstimationStatistics($A$, $(X_t, Y_t)$)
          \ENDIF
        \ENDFOR
      \ENDFOR
    \ELSIF {$I_t = \text{structure}$}
      \IF {$A_t$ has fewer than $m$ candidate split points}
        \FORALL {$d \in \operatorname{CandidateSplitDimensions}(A_t)$}
          \STATE CreateCandidateSplit($A_t$, $d, \pi_dX_t$)
        \ENDFOR
      \ENDIF
      \FORALL {$S \in \operatorname{CandidateSplits}(A_t)$}
        \FORALL {$A\in \operatorname{CandidateChildren}(S)$}
          \IF {$X_t \in A$}
            \STATE UpdateStructuralStatistics($A$, $(X_t, Y_t)$)
          \ENDIF
        \ENDFOR
      \ENDFOR
      \IF {CanSplit($A_t$)}
        \IF {ShouldSplit($A_t$)}
          \STATE Split($A_t$)
        \ELSIF {MustSplit($A_t$)}
          \STATE{Split($A_t$)}
        \ENDIF
      \ENDIF
    \ENDIF
  \ENDFOR
\end{algorithmic}
\end{algorithm}

\clearpage
\twocolumn

\begin{algorithm}
\caption{Split}
\begin{algorithmic}
\REQUIRE A leaf $A$
\REQUIRE At least one valid candidate split for exists for $A$
\STATE $S \leftarrow \operatorname{BestSplit}(A)$
\STATE $A' \leftarrow \operatorname{LeftChild}(A)$
\STATE SelectCandidateSplitDimensions($A'$,
$\min(1+\operatorname{Poisson}(\lambda), D)$) 
\STATE $A'' \leftarrow \operatorname{RightChild}(A)$
\STATE SelectCandidateSplitDimensions($A''$,
$\min(1+\operatorname{Poisson}(\lambda), D)$)
\STATE \textbf{return} $A'$, $A''$ 
\end{algorithmic}
\end{algorithm}

\begin{algorithm}
\caption{CanSplit}
\begin{algorithmic}
\REQUIRE A leaf $A$
\STATE $d \leftarrow \operatorname{Depth}(A)$
\FORALL {$S \in \operatorname{CandidateSplits}(A)$}
  \IF {SplitIsValid($A$, $S$)}
    \STATE \textbf{return} \textbf{true} 
  \ENDIF
\ENDFOR
\STATE \textbf{return} \textbf{false}
\end{algorithmic}
\end{algorithm}

\begin{algorithm}
\caption{SplitIsValid}
\begin{algorithmic}
\REQUIRE A leaf $A$
\REQUIRE A split $S$
\STATE $d \leftarrow \operatorname{Depth}(A)$
\STATE $A' \leftarrow \operatorname{LeftChild}(S)$
\STATE $A'' \leftarrow \operatorname{RightChild}(S)$
\STATE \textbf{return}  $N^e(A') \ge \alpha(d)$ and $N^e(A'') \ge \alpha(d)$
\end{algorithmic}
\end{algorithm}

\begin{algorithm}
\caption{MustSplit}
\begin{algorithmic}
\REQUIRE A leaf $A$
\STATE $d \leftarrow \operatorname{Depth}(A)$
\STATE  \textbf{return} $N^e(A) \ge \beta(d)$ 
\end{algorithmic}
\end{algorithm}

\begin{algorithm}
\caption{ShouldSplit}
\begin{algorithmic}
\REQUIRE A leaf $A$
\FORALL {$S\in \operatorname{CandidateSplits}(A)$}
\IF {$\operatorname{InformationGain}(S) > \tau$}
\IF {SplitIsValid($A$, $S$)}
\STATE \textbf{return} \textbf{true}
\ENDIF
\ENDIF
\ENDFOR
\STATE \textbf{return} \textbf{false}
\end{algorithmic}
\end{algorithm}

\begin{algorithm}
\caption{BestSplit}
\begin{algorithmic}
\REQUIRE A leaf $A$
\REQUIRE At least one valid candidate split exists for $A$
\STATE \textbf{best\_split} $\leftarrow$ \textbf{none}
\FORALL {$S \in \operatorname{CandidateSplits(A)}$}
\IF {InformationGain($A$, $S$) $\ge$ InformationGain($A$, \textbf{best\_split})}
\IF {SplitIsValid($A$, $S$)}
\STATE \textbf{best\_split} $\leftarrow$ $S$
\ENDIF
\ENDIF
\ENDFOR
\STATE \textbf{return} \textbf{best\_split}
\end{algorithmic}
\end{algorithm}

\begin{algorithm}
\caption{InformationGain}
\begin{algorithmic}
\REQUIRE A leaf $A$
\REQUIRE A split $S$
\STATE $A' \leftarrow \operatorname{LeftChild}(S)$
\STATE $A'' \leftarrow \operatorname{RightChild}(S)$
\STATE \textbf{return} $\operatorname{Entropy}(Y^s(A)) - \frac{N^s(A')}{N^s(A)}\operatorname{Entropy}(Y^s(A')) - \frac{N^s(A'')}{N^s(A)}\operatorname{Entropy}(Y^s(A''))$
\end{algorithmic}
\end{algorithm}

\begin{algorithm}
\caption{UpdateEstimationStatistics}
\begin{algorithmic}
\REQUIRE A leaf $A$
\REQUIRE A point $(X, Y)$
\STATE $N^e(A) \leftarrow N^e(A) + 1$
\STATE $Y^e(A) \leftarrow Y^e(A) + Y$
\end{algorithmic}
\end{algorithm}

\begin{algorithm}
\caption{UpdateStructuralStatistics}
\begin{algorithmic}
\REQUIRE A leaf $A$
\REQUIRE A point $(X, Y)$
\STATE $N^s(A) \leftarrow N^s(A) + 1$
\STATE $Y^s(A) \leftarrow Y^s(A) + Y$
\end{algorithmic}
\end{algorithm}


\clearpage
\onecolumn
\section{Proof of Consistency}
\label{appendix:consistency}

\subsection{A note on notation}
$A$ will be reserved for subsets of $\R^D$, and unless otherwise
indicated it can be assumed that $A$ denotes a cell of the tree
partition.  We will often be interested in the cell of the tree
partition containing a particular point, which we denote $A(x)$.
Since the partition changes over time, and therefore the shape of
$A(x)$ changes as well, we use a subscript to disambiguate: $A_t(x)$
is the cell of the partition containing $x$ at time $t$.  Cells in the
tree partition have a lifetime which begins when they are created as a
candidate child to an existing leaf and ends when they are themselves
split into two children.  When referring to a point $X_\tau \in
A_t(x)$ it is understood that $\tau$ is restricted to the lifetime of
$A_t(x)$.

We treat cells of the tree partition and leafs of the tree defining it
interchangeably, denoting both with an appropriately decorated $A$.

$N$ generally refers to the number of points of some type in some interval of
time.  A superscript always denotes type, so $N^k$ refers to a count of points
of type $k$.  Two special types, $e$ and $s$, are used to denote estimation and
structure points, respectively.  Pairs of subscripts are used to denote time
intervals, so $N_{a,b}^k$ denotes the number of points of type $k$ which appear
during the time interval $[a,b]$.  We also use $N$ as a function whose argument
is a subset of $\R^D$ in order to restrict the counting spatially:
$N_{a,b}^e(A)$ refers to the number of estimation points which fall in the set
$A$ during the time interval $[a,b]$.  We make use of one additional variant of
$N$ as a function when its argument is a cell in the partition: when we write
$N^k(A_t(x))$, without subscripts on $N$, the interval of time we count over is
understood to be the lifetime of the cell $A_t(x)$.

\subsection{Preliminaries}

\begin{lemma}
  Suppose we partition a stream of data into $c$ parts by assigning each point
  $(X_t, Y_t)$ to part $I_t \in \{1, \ldots, c\}$ with fixed probability
  $p_k$, meaning that
  \begin{align}
    N^k_{a,b} = \sum_{t=a}^b \1{I_t = k}
    \enspace.
    \label{eq:Ntk}
  \end{align}
  Then with probability 1, $N_{a,b}^k \to \infty$ for all $k\in \{1,
  \ldots, c\}$ as $b-a\to\infty$.
  \label{lemma:big-parts}
\end{lemma}
\begin{proof}
  Note that $\P{I_t = 1} = p_1$ and these events are independent for each $t$.
  By the second Borel-Cantelli lemma, the probability that the events in this
  sequence occur infinitely often is 1.  The cases for $I_t \in \{2, \ldots,
  c\}$ are similar.
\end{proof}

\begin{lemma}
  Let $X_t$ be a sequence of iid random variables with distribution
  $\mu$, let $A$ be a fixed set such that $\mu(A) > 0$ and let
  $\{I_t\}$ be a fixed partitioning sequence.  Then the random
  variable
  \begin{align*}
    N_{a,b}^k(A) = \sum_{a\le t\le b:I_t=k}\1{X_t \in A}
  \end{align*}
  is Binomial with parameters $N_{a,b}^k$ and $\mu(A)$.  In particular,
  \begin{align*}
    \P{N_{a,b}^k(A) \le \frac{\mu(A)}{2}N_{a,b}^k} \le \exp{-\frac{\mu(A)^2}{2}N_{a,b}^k}
  \end{align*}
  which goes to 0 as $b-a\to\infty$, where $N_{a,b}^k$ is the
  deterministic quantity defined as in Equation~\ref{eq:Ntk}.
  \label{lemma:binomial}
\end{lemma}
\begin{proof}
  $N_{a,b}^k(A)$ is a sum of iid indicator random variables so it is
  Binomial.  It has the specified parameters because it is a sum over
  $N_{a,b}^k$ elements and $\P{X_t \in A} = \mu(A)$.  Moreover,
  $\E{N_{a,b}^k(A)} = \mu(A)N_{a,b}^k$ so by Hoeffding's inequality we have
  that
  \begin{align*}
    \P{N_{a,b}^k(A) \le \E{N_{a,b}^k(A)} - \epsilon N_{a,b}^k} = \P{N_{a,b}^k(A)
      \le N_{a,b}^k(\mu(A) - \epsilon)} \le \exp{-2\epsilon^2 N_{a,b}^k}
    \enspace.
  \end{align*}
  Setting $\epsilon = \frac{1}{2}\mu(A)$ gives theresult.
\end{proof}

\subsection{Proof of Proposition~\ref{prop:multi-biau}}
\begin{proof}
  Let $g(x)$ denote the Bayes classifier.  Consistency of $\{g_t\}$ is
  equivalent to saying that $\E{L(g_t)} = \P{g_t(X, Z) \neq Y} \to L^*$.  In
  fact, since $\P{g_t(X,Z)\neq Y\,|\,X=x} \ge \P{g(X)\neq Y\,|\, X=x}$ for all
  $x \in \R^D$, consistency of $\{g_t\}$ means that for $\mu$-almost all $x$,
  \begin{align*}
    \P{g_t(X,Z)\neq Y\,|\, X=x} \to \P{g(X)\neq Y\,|\,X=x} = 1 -
    \max_k\{\eta^k(x)\}
  \end{align*}
  Define the following two sets of indices
  \begin{align*}
    G &= \{ k \,|\, \eta^k(x) = \max_k\{\eta^k(x)\} \} \enspace,
    \\
    B &= \{ k \,|\, \eta^k(x) < \max_k\{\eta^k(x)\} \} \enspace.
  \end{align*}
  Then
  \begin{align*}
    \P{g_t(X,Z)\neq Y\,|\, X=x} &= \sum_k \P{g_t(X,Z) = k\,|\, X=x}\P{Y\neq
      k|X=x}
    \\
    &\le (1-\max_k\{\eta^k(x)\})\sum_{k\in G} \P{g_t(X,Z) = k\,|\, X=x} +
    \sum_{k\in B} \P{g_t(X,Z) = k\,|\, X=x}
    \enspace,
  \end{align*}
  which means it suffices to show that $\P{g_t^{(M)}(X,Z^M)=k\,|\,X=x}
  \to 0$ for all $k \in B$.  However, using $Z^M$ to denote $M$
  (possibly dependent) copies of $Z$, for all $k\in B$ we have
  \begin{align*}
    \P{g_t^{(M)}(x, Z^M) = k} &= \P{\sum_{j=1}^M\1{g_t(x, Z_j) = k} >
      \max_{c\neq k} \sum_{j=1}^M\1{g_t(x, Z_j) = c}}
    \\
    &\le \P{\sum_{j=1}^M\1{g_t(x, Z_j) = k} \ge 1}
    \intertext{By Markov's inequality,} &\le \E{\sum_{j=1}^M\1{g_t(x, Z_j) = k}}
    \\
    &= M\P{g_t(x,Z)=k} \to 0
  \end{align*}
\end{proof}

\subsection{Proof of Proposition~\ref{prop:condition-on-full-measure}}

\begin{proof}
  The sequence in question is uniformly integrable, so it is
  sufficient to show that $\E{\P{g_t(X,Z,I)\neq Y\,|\,I}} \to L^*$
  implies the result, where the expectation is taken over the random
  selection of training set.

  We can write
  \begin{align*}
    \P{g_t(X,Z,I)\neq Y} &= \E{\P{g_t(X,Z,I)\neq Y\,|\,I}}
    \\
    &= \int_{\mathcal{I}}\P{g_t(X,Z,I)\neq Y\,|\,I}\nu(I) +
    \int_{\mathcal{I}^c}\P{g_t(X,Z,I)\neq Y\,|\,I}\nu(I)
  \end{align*}
  By assumption $\nu(\mathcal{I}^c)=0$, so we have
  \begin{align*}
    \lim_{t\to\infty} \P{g_t(X,Z,I)\neq Y} &=
    \lim_{t\to\infty} \int_{\mathcal{I}} \P{g_t(X,Z,I)\neq
      Y\,|\,I}\nu(I)
    \intertext{Since probabilities are bounded in the interval $[0,
      1]$, the dominated convergence theorem allows us to exchange the
      integral and the limit,} &= \int_{\mathcal{I}}
    \lim_{t\to\infty}\P{g_t(X,Z,I)\neq Y\,|\,I}\nu(I)
    \intertext{and by assumption the conditional risk converges to the
      Bayes risk for all $I\in\mathcal{I}$, so} &=
    L^*\int_{\mathcal{I}}\nu(I)
    \\
    &= L^*
  \end{align*}
  which proves the claim.
\end{proof}

\subsection{Proof of Proposition~\ref{prop:multiclass}}
\begin{proof}
  By definition, the rule
  \begin{align*}
    g(x) &= \arg\max_k\{\eta^k(x)\}
  \end{align*}
  (where ties are broken in favour of smaller $k$) achieves the Bayes
  risk.  In the case where all the $\eta^k(x)$ are equal there is
  nothing to prove, since all choices have the same probability of
  error.  Therefore, suppose there is at least one $k$ such that
  $\eta^k(x) < \eta^{g(x)}(x)$ and define
  \begin{align*}
    m(x) &= \eta^{g(x)}(x) - \max_{k}\{\eta^k(x) \,|\, \eta^k(x) <
    \eta^{g(x)}(x) \}
    \\
    m_t(x) &= \eta_t^{g(x)}(x) - \max_{k}\{\eta_t^k(x) \,|\, \eta^k(x) <
    \eta^{g(x)}(x) \}
  \end{align*}
  The function $m(x) \ge 0$ is the margin function which measures how much
  better the best choice is than the second best choice, ignoring possible ties
  for best.  The function $m_t(x)$ measures the margin of $g_t(x)$.  If $m_t(x)
  > 0$ then $g_t(x)$ has the same probability of error as the Bayes classifier.

  The assumption above guarantees that there is some $\epsilon$ such that $m(x)
  > \epsilon$.  Using $C$ to denote the number of classes, by making $t$ large
  we can satisfy
  \begin{align*}
    \P{|\eta_t^k(X)-\eta^k(X)| < \epsilon/2} \ge 1-\delta/C
  \end{align*}
  since $\eta^k_t$ is consistent.  Thus
  \begin{align*}
    \P{\bigcap_{k=1}^C|\eta_t^k(X) - \eta^k(X)| < \epsilon/2} &\ge
    1 - K + \sum_{k=1}^C \P{|\eta_t^k(X)-\eta^k(X)| < \epsilon/2 }
    \ge 1-\delta
  \end{align*}
  So with probability at least $1-\delta$ we have
  \begin{align*}
    m_t(X) &= \eta_t^{g(X)} - \max_{k}\{\eta_t^k(X) \,|\, \eta^k(X) <
    \eta^{g(X)}(X) \}
    \\
    &\ge (\eta^{g(X)} - \epsilon/2) - \max_{k}\{\eta_t^k(X) + \epsilon/2 \,|\,
    \eta^k(X) < \eta^{g(x)}(X) \}
    \\
    &= \eta^{g(X)} - \max_{k}\{\eta^k(X) \,|\, \eta^k(X) <
    \eta^{g(X)}(X) \} - \epsilon
    \\
    &= m(X) - \epsilon
    \\
    &> 0
  \end{align*}
  Since $\delta$ is arbitrary this means that the risk of $g_t$
  converges in probability to the Bayes risk.
\end{proof}

\subsection{Proof of Theorem~\ref{thm:consistent-forest}}

The proof of Theorem~\ref{thm:consistent-forest} is built in several
pieces.

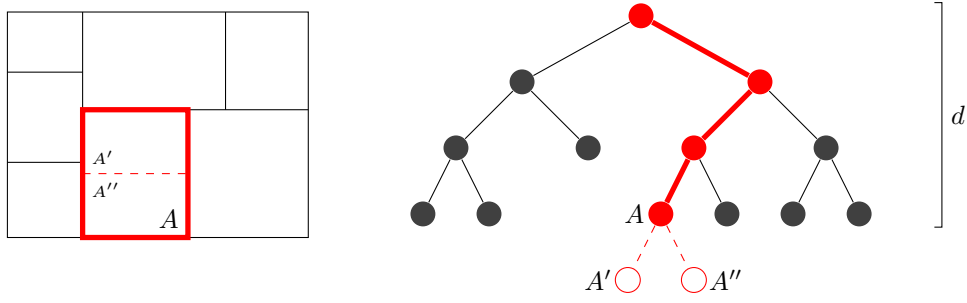
\begin{figure}
  \centering
  \raisebox{27pt}{
  \begin{tikzpicture}
    \draw (0, 0) rectangle (4,3);
    \draw (1, 0) -- (1, 3);
    \draw (1, 1.7) -- (4, 1.7);
    \draw (2.9, 1.7) -- (2.9, 3);
    \draw (2.4, 1.7) -- (2.4,0);
    \draw (0, 1) -- (1, 1);
    \draw (0, 2.2) -- (1, 2.2);

    \draw[line width=2, color=red] (1,0) rectangle (2.4,1.7);
    \draw (2.4, 0) node [above left] {$A$};
    \draw[dashed, color=red] (1, 0.85) -- (2.4, 0.85);
    \draw (1, 0.85)
    node [above right] {\tiny{$A'$}}
    node [below right] {\tiny{$A''$}};

  \end{tikzpicture}
  \hspace{1cm}
  }
  \begin{tikzpicture}
    [
    level distance=25,
    level 1/.style={sibling distance=90},
    level 2/.style={sibling distance=50},
    level 3/.style={sibling distance=25},
    emph/.style={fill=red, circle, edge from parent/.style={red,line width=2,draw}},
    norm/.style={fill=black!75,circle, edge from parent/.style={black,thin,draw}}
    ]
    \node [emph] (root) {}
    child {node [norm] {}
      child {node [norm] {}
        child {node [norm] {}}
        child {node [norm] {}}
      }
      child {node [norm] {}}
    }
    child[emph] {node [emph] {}
      child[emph] {node [emph] {}
        child[emph] {   
          node [emph] (A) {}
          child [edge from parent/.style={red,thin,dashed,draw}] {
            node (Aprime) [circle, thin, draw] {}
            node [left] {$A'$}
          }
          child [edge from parent/.style={red,thin,dashed,draw}] {
            node [circle, thin, draw] {}
            node [right] {$A''$}
          }
          node [left] {$A$}
        }
        child[norm] {node [norm] {}}
      }
      child[norm] {node [norm] {}
        child {node [norm] {}}
        child {node [norm] {}}
      }
    };

    \path let \p1 = (root) in node (topright) at (4,\y1 + 5) {};
    \path let \p1 = (A) in node (bottomright) at (4,\y1 - 5) {};

    \draw
    (topright.center)
    ++ (-0.1, 0)
    -- (topright.center)
    -- (bottomright.center)
    -- ++(-0.1, 0);
    \draw ($ (topright) !.5! (bottomright) $) node [right] {$d$};
  \end{tikzpicture}
  \caption{This Figure shows the setting of
    Proposition~\ref{prop:will-split}.  Conditioned on a partially
    built tree we select an arbitrary leaf at depth $d$ and an
    arbitrary candidate split in that leaf.  The proposition shows
    that, assuming no other split for $A$ is selected, we can
    guarantee that the chosen candidate split will occur in bounded
    time with arbitrarily high probability.}
  \label{fig:will-split}
\end{figure}

\begin{proposition}
  Fix a partitioning sequence.  Let $t_0$ be a time at which a split
  occurs in a tree built using this sequence, and let $g_{t_0}$ denote
  the tree after this split has been made.  If $A$ is one of the newly
  created cells in $g_{t_0}$ then we can guarantee that the cell $A$
  is split before time $t > t_0$ with probability at least $1-\delta$
  by making $t$ sufficiently large.
  \label{prop:will-split}
\end{proposition}
\begin{proof}
  Let $d$ denote the depth of $A$ in the tree $g_{t_0}$ and note that
  $\mu(A) > 0$ with probability 1 since $X$ has a density.  This
  situation is illustrated in Figure~\ref{fig:will-split}.  By
  construction, if the following conditions hold:
  \begin{enumerate}
  \item For some candidate split in $A$, the number of estimation
    points in both children is at least $\alpha(d)$,
  \item The number of estimation points in $A$ is at least
    $\beta(d)$,
  \end{enumerate}
  then the algorithm must split $A$ when the next structure point
  arrives.  Thus in order to force a split we need the following
  sequence of events to occur:
  \begin{enumerate}
  \item A structure point must arrive in $A$ to create a candidate
    split point.
  \item The above two conditions must be satisfied.
  \item Another structure point must arrive in $A$ to force a split.
  \end{enumerate}
  It is possible for a split to be made before these events occur, but
  assuming a split is not triggered by some other mechanism we can
  guarantee that this sequence of events will occur in bounded time
  with high probability.

  Suppose a split is not triggered by a different mechanism.  Define
  $E_0$ to be an event that occurs at $t_0$ with probability 1, and
  let $E_1 \le E_2 \le E_3$ be the times at which the above numbered
  events occur.  Each of these events requires the previous one to
  have occurred and moreover, the sequence has a Markov structure, so
  for $t_0 \le t_1 \le t_2 \le t_3 = t$ we have
  \begin{align*}
    \P{E_1 \le t \cap E_2 \le t \cap E_3 \le t\,|\,E_0=t_0} &\ge \P{E_1 \le t_1
      \cap E_2 \le t_2 \cap E_3 \le t_3\,|\,E_0=t_0}
    \\
    &= \P{E_1 \le t_1\,|\,E_0=t_0}\P{E_2 \le t_2\,|\,E_1 \le t_1}\P{ E_3 \le
      t_3\,|\,E_2 \le t_2}
    \\
    &\ge \P{E_1 \le t_1\,|\,E_0=t_0}\P{E_2 \le t_2\,|\,E_1 = t_1}\P{ E_3 \le
      t_3\,|\,E_2 = t_2}
    \enspace.
  \end{align*}
  We can rewrite the first and last term in more friendly notation as
  \begin{align*}
    \P{E_1 \le t_1\,|\,E_0=t_0} &= \P{N_{t_0,t_1}^s(A) \ge 1} \enspace,
    \\
    \P{E_3 \le t_3 \,|\, E_2 = t_2} &= \P{N_{t_2,t_3}^s(A) \ge 1}
    \enspace.
  \end{align*}
  Lemma~\ref{lemma:binomial} allows us to lower bound both of these
  probabilities by $1-\epsilon$ for any $\epsilon>0$ by making
  $t_1-t_0$ and $t_3-t_2$ large enough that
  \begin{align*}
    N_{t_0,t_1}^s &\ge \frac{2}{\mu(A)}\max\left\{1 ,\,
      \mu(A)^{-1}\log\left(\frac{1}{\epsilon}\right) \right\}
    \intertext{and} N_{t_2,t_3}^s &\ge \frac{2}{\mu(A)}\max\left\{1
      ,\, \mu(A)^{-1}\log\left(\frac{1}{\epsilon}\right) \right\}
  \end{align*}
  respectively.  To bound the centre term, recall that $\mu(A')>0$ and
  $\mu(A'')>0$ with probability 1, and $\beta(d) \ge \alpha(d)$ so
  \begin{align*}
    \P{E_2 \le t_2 \,|\,E_1=t_1} &\ge \P{N_{t_1,t_2}^e(A') \ge \beta(d)
      \cap N_{t_1,t_2}^e(A'') \ge \beta(d)}
    \\
    &\ge \P{N_{t_1,t_2}^e(A') \ge \beta(d)} + \P{N_{t_1,t_2}^e(A'') \ge
      \beta(d)} - 1 \enspace,
  \end{align*}
  and we can again use Lemma~\ref{lemma:binomial} lower bound this by
  $1-\epsilon$ by making $t_2 - t_1$ sufficiently large that
  \begin{align*}
    N_{t_1,t_2}^e &\ge \frac{2}{\min\{\mu(A'), \mu(A'')\}}
    \max\left\{\beta(d),\, \min\{\mu(A'), \mu(A'')\}^{-1}
      \log\left(\frac{2}{\epsilon}\right) \right\}
  \end{align*}

  Thus by setting $\epsilon = 1-(1-\delta)^{1/3}$ can ensure that the
  probability of a split before time $t$ is at least $1-\delta$ if we
  make
  \begin{align*}
    t = t_0 + (t_1 - t_0) + (t_2 - t_1) + (t_3 - t_2)
  \end{align*}
  sufficiently large.
\end{proof}

\begin{figure}
  \centering
  \begin{tikzpicture}
    \draw [->] (-1, 1) -- (11, 1) node [right] {$t$};
    \draw (0, 1) node [below] {$t_0$};
    \draw (0, 1) -- ++ (0, 0.1) node [above] {$E_0$};
    \draw (3, 1) -- ++ (0, 0.1) node [above] {$E_1$};
    \draw (7, 1) -- ++ (0, 0.1) node [above] {$E_2$};
    \draw (10, 1) -- ++ (0, 0.1) node [above] {$E_3$};

    \def\interval[#1]{++ (0, 0.1) -- ++ (0, -0.1) -- 
      node[above, near start]{#1}
      ++ (4.5, 0) -- ++ (0, 0.1)}

    \draw (0, 0.0) \interval [$t_1-t_0$];
    \draw (3, 0.25) \interval[$t_2-t_1$];
    \draw (7, 0.5) \interval[$t_3-t_2$];

  \end{tikzpicture}
  \caption{This Figure diagrams the structure of the argument used in
    Propositions~\ref{prop:will-split}
    and~\ref{prop:infinite-splits}.  The indicated intervals are show
    regions where the next event must occur with high probability.
    Each of these intervals is finite, so their sum is also finite.
    We find an interval which contains all of the events with high
    probability by summing the lengths of the intervals for which we
    have individual bounds.}
\end{figure}
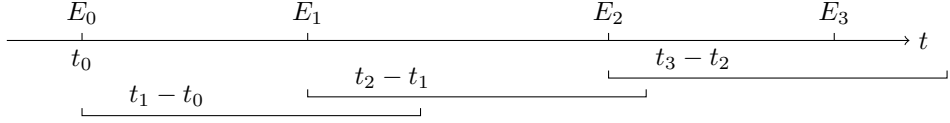

\begin{proposition}
  Fix a partitioning sequence.  Each cell in a tree built based on this sequence
  is split infinitely often in probability.  i.e all $K>0$ and any $x$ in the
  support of $X$,
  \begin{align*}
    \P{A_t(x) \text{ has been split fewer than $K$ times}} \to 0
  \end{align*}
  as $t\to\infty$.
  \label{prop:infinite-splits}
\end{proposition}
\begin{proof}
  For an arbitrary point $x$ in the support of $X$, let $E_k$ denote
  the time at which the cell containing $x$ is split for the $k$th
  time, or infinity if the cell containing $x$ is split fewer than $k$
  times (define $E_0 = 0$ with probability 1). Now define the
  following sequence:
  \begin{align*}
    t_0 &= 0 \\
    t_i &= \min\{t\,|\, \P{E_i \le t \,|\, E_{i-1} = t_{i-1}} \ge
    1-\epsilon \}
  \end{align*}
  and set $T_\delta = t_k$. Proposition~\ref{prop:will-split} guarantees
  that each of the above $t_i$'s exists and is finite.  Compute,
  \begin{align*}
    \P{E_k \le T_\delta} &= \P{\bigcap_{i=1}^k [E_i \le T_\delta]}
    \\
    &\ge \P{\bigcap_{i=1}^k [E_i \le t_i]}
    \\
    &= \prod_{i=1}^k\P{E_i \le t_i \,|\, \bigcap_{j<i} [E_j \le t_j]}
    \\
    &= \prod_{i=1}^k\P{E_i \le t_i \,|\, E_{i-1} \le t_{i-1}}
    \\
    &\ge \prod_{i=1}^k \P{E_i \le t_i \,|\, E_{i-1} = t_{i-1}}
    \\
    &\ge (1-\epsilon)^k
  \end{align*}
  where the last line follows from the choice of $t_i$'s.  Thus for any $\delta
  > 0$ we can choose $T_\delta$ to guarantee $\P{E_k \le T_\delta} \ge 1 -
  \delta$ by setting $\epsilon = 1-(1-\delta)^{1/k}$ and applying the above
  process.  We can make this guarantee for any $k$ which allows us to conclude
  that $\P{E_k \le t} \to 1$ as $t\to\infty$ for all $k$ as required.
\end{proof}

\begin{proposition}
  Fix a partitioning sequence.  Let $A_t(X)$ denote the cell of $g_t$
  (built based on the partitioning sequence) containing the point $X$.
  Then $\diam(A_t(X)) \to 0$ in probability as $t\to\infty$.
  \label{prop:diam-to-zero}
\end{proposition}
\begin{proof}
  Let $V_t(x)$ be the size of the first dimension of $A_t(x)$.  It
  suffices to show that $\E{V_t(x)} \to 0$ for all $x$ in the support
  of $X$.

  Let $X_1, \ldots, X_{m'} \sim \mu|_{A_t(x)}$ for some $1 \le m' \le
  m$ denote the samples from the structure stream that are used to
  determine the candidate splits in the cell $A_t(x)$.  Use $\pi_d$ to
  denote a projection onto the $d$th coordinate, and without loss of
  generality, assume that $V_t = 1$ and $\pi_1 X_i \sim
  \operatorname{Uniform}[0, 1]$.  Conditioned on the event that the
  first dimension is cut, the largest possible size of the first
  dimension of a child cell is bounded by
  \begin{align*}
    V^* = \max(\max_{i=1}^m\pi_1X_i, 1-\min_{i=1}^m\pi_1X_i) \enspace.
  \end{align*}
  Recall that we choose the number of candidate dimensions as
  $\min(1+\operatorname{Poisson}(\lambda), D)$ and select that number
  of distinct dimensions uniformly at random to be candidates.  Define
  the following events:
  \begin{align*}
    E_1 &= \{ \text{There is exactly one candidate dimension} \}
    \\
    E_2 &= \{ \text{The first dimension is a candidate} \}
  \end{align*}
  Then using $V'$ to denote the size of the first dimension of the
  child cell,
  \begin{align*}
    \E{V'} &\le \E{\1{(E_1 \cap E_2)^c} + \1{E_1 \cap E_2}V^*}
    \\
    &= \P{E_1^c} + \P{E_2^c | E_1}\P{E_1} + \P{E_2 |
      E_1}\P{E_1}\E{V^*}
    \\
    &= (1-e^{-\lambda}) + (1-\frac{1}{d})e^{-\lambda} +
    \frac{1}{d}e^{-\lambda}\E{V^*}
    \\
    &= 1 - \frac{e^{-\lambda}}{D} + \frac{e^{-\lambda}}{D}\E{V^*}
    \\
    &= 1 - \frac{e^{-\lambda}}{D} +
    \frac{e^{-\lambda}}{D}\E{\max(\max_{i=1}^m\pi_1X_i,
      1-\min_{i=1}^m\pi_1X_i)}
    \\
    &= 1 - \frac{e^{-\lambda}}{D} +
    \frac{e^{-\lambda}}{D}\cdot\frac{2m+1}{2m+2}
    \\
    &= 1 - \frac{e^{-\lambda}}{2D(m+1)}
  \end{align*}
  Iterating this argument we have that after $K$ splits the expected
  size of the first dimension of the cell containing $x$ is upper
  bounded by
  \begin{align*}
    \left(1 - \frac{e^{-\lambda}}{2D(m+1)}\right)^K
  \end{align*}
  so it suffices to have $K\to\infty$ in probability, which we know to
  be the case from Proposition~\ref{prop:infinite-splits}.
\end{proof}

\begin{proposition}
  Fix a partitioning sequence.  In any tree built based on this
  sequence, $N^e(A_t(X)) \to \infty$ in probability.
  \label{prop:n-to-infty}
\end{proposition}
\begin{proof}
  It suffices to show that $N^e(A_t(x)) \to \infty$ for all $x$ in the support
  of $X$.  Fix such an $x$, by Proposition~\ref{prop:infinite-splits} we
  can make the probability $A_t(x)$ is split fewer than $K$ times
  arbitrarily small for any $K$.  Moreover, by construction
  immediately after the $K$-th split is made the number of estimation
  points contributing to the prediction at $x$ is at least $\alpha(K)$, and
  this number can only increase.  Thus for all $K$ we have that
  $\P{N^e(A_t(x)) < \alpha(K)} \to 0$ as $t\to\infty$ as required.
\end{proof}

We are now ready to prove our main result.  All the work has been
done, it is simply a matter of assembling the pieces.

\begin{proof}[Proof (of Theorem~\ref{thm:consistent-forest})]
  Fix a partitioning sequence.  Conditioned on this sequence the consistency of
  each of the class posteriors follows from Theorem~\ref{thm:devroye61}.  The
  two required conditions where shown to hold in
  Propositions~\ref{prop:diam-to-zero} and~\ref{prop:n-to-infty}.  Consistency
  of the multiclass tree classifier then follows by applying
  Proposition~\ref{prop:multiclass}.

  To remove the conditioning on the partitioning sequence, note that
  Lemma~\ref{lemma:big-parts} shows that our tree generation mechanism produces
  a partitioning sequence with probability 1.  Apply
  Proposition~\ref{prop:condition-on-full-measure} to get unconditional
  consistency of the multiclass tree.

  Proposition~\ref{prop:multi-biau} lifts consistency of the trees to
  consistency of the forest, establishing the desired result.
\end{proof}

\subsection{Extension to a Fixed Size Fringe}

Proving consistency is preserved with a fixed size fringe requires more precise
control over the relationship between the number of estimation points seen in an
interval, $N_{t_0,t}^e$, and the total number of splits which have occurred in
the tree, $K$.  The following two lemmas provide the control we need.
\begin{lemma}
  Fix a partitioning sequence.  If $K$ is the number of splits which have
  occurred at or before time $t$ then for all $M>0$
  \begin{align*}
    \P{K \le M} \to 0
  \end{align*}
  in probability as $t\to\infty$.
\end{lemma}
\begin{proof}
  Denote the fringe at time $t$ with $F_t$ which has max size $|F|$, and the set
  of leafs at time $t$ as $L_t$ with size $|L_t|$.  If $|L_t| < |F|$ then there
  is no change from the unbounded fringe case, so we assume that $|L_t| \ge
  |F|$ so that for all $t$ there are exactly $|F|$ leafs in the fringe.

  Suppose a leaf $A_1 \in F_{t_0}$ for some $t_0$ then for every $\delta > 0$
  there is a finite time $t_1$ such that for all $t\ge t_1$
  \begin{align*}
    \P{A_1\text{ has not been split before time }t} \le \frac{\delta}{|F|}
  \end{align*}

  Now fix a time $t_0$ and $\delta>0$.  For each leaf $A_i \in F_{t_0}$ we can
  choose $t_i$ to satisfy the above bound.  Set $t = \max_{i} t_i$ then the
  union bound gives
  \begin{align*}
    \P{K \le |F| \text{ at time } t} \le \delta
  \end{align*}
  Iterate this argument $\ceil{M/|F|}$ times with $\delta =
  \epsilon/\ceil{M/|F|}$ and apply the union bound again to get that for
  sufficiently large $t$
  \begin{align*}
    \P{K\le M} \le \epsilon
  \end{align*}
  for any $\epsilon > 0$.
\end{proof}

\begin{lemma}
  Fix a partitioning sequence.  If $K$ is the number of splits which have
  occurred at or before time $t$ then for any $t_0 > 0$, $K/N_{t_0,t}^e \to 0$
  as $t\to\infty$.
  \label{lemma:upper-bound-K}
\end{lemma}
\begin{proof}
  First note that $N_{t_0,t}^e = N_{0,t}^e - N_{0,t_0-1}^e$ so
  \begin{align*}
    \frac{K}{N_{t_0,t}^e} = \frac{K}{N_{0,t}^e - N_{0,t_0-1}^e}
  \end{align*}
  and since $N_{0,t_0-1}^e$ is fixed it is sufficient to show that
  $K/N_{0,t}^e\to0$.  In the following we write $N=N_{0,t}^e$ to lighten the
  notation.

  Define the cost of a tree $T$ as the minimum value of $N$ required to
  construct a tree with the same shape as $T$.  The cost of the tree is governed
  by the function $\alpha(d)$ which gives the cost of splitting a leaf at level
  $d$.  The cost of a tree is found by summing the cost of each split required
  to build the tree.

  Note that no tree on $K$ splits is cheaper than a tree of max depth $d =
  \ceil{\log_2(K)}$ with all levels full (except possibly the last, which may be
  partially full).  This is simple to see, since $\alpha(d)$ is an increasing
  function of $d$, meaning it is never more expensive to add a node at a lower
  level than a higher one.  Thus we assume wlog that the tree is full except
  possibly in the last level.



  When filling level $d$ of the tree, each split incurs a cost of at least
  $2\alpha(d+1)$ points.  This also tells us that filling level $d$ requires
  that $N$ increase by at least $2^d\alpha(d)$ (filling level $d$ corresponds to
  splitting each of the $2^{d-1}$ leafs on level $d-1$).  Filling the first $d$
  levels incurs a cost of at least
  \begin{align*}
    N_d = \sum_{k=1}^{d}2^k\alpha(k)
  \end{align*}
  points.  When $N=N_d$ the tree can be at most a full binary tree of depth
  $d$, meaning that $K \le 2^{d}-1$.

  The above argument gives a collection of linear upper bounds on $K$ in terms
  of $N$.  We know that the maximum growth rate is linear between $(N_d, 2^d-1)$
  and $(N_{d+1}, 2^{d+1}-1)$ so for all $d$ we can find that since
  \begin{align*}
    \frac{(2^{d+1}-1) - (2^{d}-1)}{(N_{d+1}) - (N_{d})} = \frac{2^{d+1} -
      2^{d}}{\sum_{k=1}^{d+1}2^k\alpha(k) - \sum_{k=1}^{d}2^k\alpha(k)}
    = \frac{2^{d}}{2^{d+1}\alpha(d+1)} = \frac{1}{2\alpha(d+1)}
  \end{align*}
  we have that for all $N$ and $d$,
  \begin{align*}
    K \le \frac{1}{2\alpha(d+1)}N + C(d)
  \end{align*}
  where $C(d)$ is given by
  \begin{align*}
    C(d) &= 2^d - 1 - \frac{1}{2}\sum_{k=1}^{d}2^k\frac{\alpha(k)}{\alpha(d+1)}
    \enspace.
  \end{align*}
  From this we have
  \begin{align*}
    \frac{K}{N} \le \frac{1}{2\alpha(d+1)} + \frac{1}{N}\left( 2^d - 1 -
      \frac{1}{2}\sum_{k=1}^{d}2^k\frac{\alpha(k)}{\alpha(d+1)} \right)
    \enspace,
  \end{align*}
  so if we choose $d$ to make $1/\alpha(d+1) \le \delta/2$ and then pick $N$
  such that $C(d)/N \le \delta/2$ we have $K/N \le \delta$ for arbitrary $\delta
  > 0$ which proves the claim.
\end{proof}

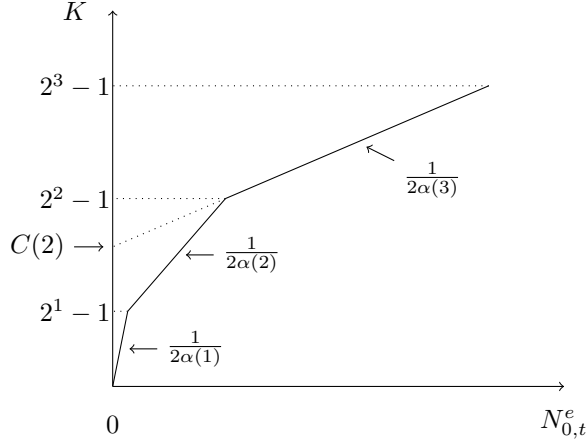
\begin{figure}[h]
  \centering
  
  \begin{tikzpicture}
    \draw [->] (0,0) -- (6,0);
    \draw [->] (0,0) -- (0, 5);
    
    \node (hbaseline) at (0,-0.5) {};
    \node (vbaseline) at (-0.5, 0) {};
    \node (ylabel) at (vbaseline |- 0,5) {$K$};
    \node (xlabel) at (hbaseline -| 6,0) {$N_{0,t}^e$};

    \node (origin) at (0,0) {};
    \node (1) at (0.2, 1) {};
    \node (2) at (1.5, 2.5) {};
    \node (3) at (5, 4) {};

    \draw (origin.center) -- (1.center);
    \draw (1.center) -- (2.center);
    \draw (2.center) -- (3.center);
    
    \node () at (origin |- hbaseline) {$0$};
    
    \node () at (1 -| vbaseline) {$2^1-1$};
    \draw [dotted] (1.center) -- (1 -| origin);
    
    \node () at (2 -| vbaseline) {$2^2-1$};
    \draw [dotted] (2.center) -- (2 -| origin);
    
    \node () at (3 -| vbaseline) {$2^3-1$};
    \draw [dotted] (3.center) -- (3 -| origin);

    \path [name path=yinterceptextention] (2.center) --
    ($(2.center)!(origin.center)!(3.center)$);
    \path [name path=yaxis] (origin.center) --++ (0,5);
    \node [name intersections={of=yinterceptextention and yaxis}](yintercept)
    at (intersection-1) {};
    \draw [dotted] (2.center) -- (yintercept.center);
    \node (yinterceptlabel) at ($(yintercept.center) + (-1,0)$) {$C(2)$};
    \draw [->] (yinterceptlabel) -- (yintercept);

    \node (label1head) at ($(1) !.5! (origin)$) {};
    \node (label1tag) at ($(label1head) + (1,0)$) {$\frac{1}{2\alpha(1)}$};
    \draw [->] (label1tag) -- (label1head);
    
    \node (label2head) at ($(2) !.5! (1)$) {};
    \node (label2tag) at ($(label2head) + (1, 0)$) {$\frac{1}{2\alpha(2)}$};
    \draw [->] (label2tag) -- (label2head);
    
    \node (label3head) at ($(3) !.5! (2)$) {};
    \node (label3tag) at ($(label3head) + (1, -0.5)$) {$\frac{1}{2\alpha(3)}$};
    \draw [->] (label3tag) -- (label3head);
  \end{tikzpicture}

  \caption{Diagram of the bound in Lemma~\ref{lemma:upper-bound-K}.  The
    horizontal axis is the number of estimation points seen at time $t$ and the
    vertical axis is the number of splits.  The first bend is the earliest point
    at which the root of the tree could be split, which requires $2\alpha(1)$
    points to create 2 new leafs at level $1$.  Similarly, the second bend is
    the point at which all leafs at level 1 have been split, each of which
    requires at least $2\alpha(2)$ points to create a pair of leafs at level
    $2$.}
\end{figure}

In order to show that our algorithm remains consistent with a fixed size fringe
we must ensure that Proposition~\ref{prop:will-split} does not fail in this
setting.  Interpreted in the context of a finite fringe,
Proposition~\ref{prop:will-split} says that any cell in the fringe will be split
in finite time.  This means that to ensure consistency we need only show that
any inactive point will be added to the fringe in finite time.

\begin{remark}
  If $s(A) = 0$ for any leaf then we know that $e(A) = 0$, since $\mu(A) > 0$ by
  construction.  If $e(A) = 0$ then $\P{g(X) \neq Y\,|\,X\in A} = 0$ which means
  that any subdivision of $A$ has the same asymptotic probability of error as
  leaving $A$ in tact.  Our rule never splits $A$ and thus fails to satisfy the
  shrinking leaf condition, but our predictions are asymptotically the same as
  if we had divided $A$ into arbitrarily many pieces so this doesn't matter.
\end{remark}

\begin{proposition}
  Every leaf with $s(A) > 0$ will be added to the fringe in finite time with
  arbitrarily high probability.
\end{proposition}
\begin{proof}
  Pick an arbitrary time $t_0$ and condition on everything before $t_0$.  For an
  arbitrary node $A \subset \R^D$, if $A'$ is a child of $A$ then we know that
  if $\{U_i\}_{i=1}^{Dm}$ are iid on $[0,1]$ then
  \begin{align*}
    \E{\mu(A')} &\le \mu(A)\E{\max_{i=1}^{Dm} (\max(U_i, 1-U_i))}
    \\
    &= \mu(A)\left(\frac{2Dm+1}{2Dm+2}\right)
  \end{align*}
  since there are at most $D$ candidate dimensions and each one accumulates at
  most $m$ candidate splits.  So if $A^K$ is any leaf created by $K$ splits of
  $A$ then
  \begin{align*}
    \E{\mu(A^K)} \le \mu(A)\left(\frac{2Dm+1}{2Dm+2}\right)^K
  \end{align*}
  Notice that since we have conditioned on the tree at $t_0$ so,
  \begin{align*}
    \E{\hat{p}(A^K)} = \E{\E{\hat{p}(A^K)\,|\,\mu(A^K)}} = \E{\mu(A^K)}
    \enspace.
  \end{align*}
  We can bound $\hat{p}(A^K)$ with Hoeffding's inequality,
  \begin{align*}
    \P{\hat{p}(A^K) \ge \mu(A)\left(\frac{2Dm+1}{2Dm+2}\right)^K + \epsilon} &\le
    \exp{-2|A^K|\epsilon^2}
    \enspace.
  \end{align*}
  Set $(2^{K+1}|L|)^{-1}\delta = \exp{-2|A^K|\epsilon^2}$ and invert the bound
  to get
  \begin{align*}
    \P{\hat{p}(A^K) \ge \mu(A)\left(\frac{2Dm+1}{2Dm+2}\right)^K +
      \sqrt{\frac{1}{2|A^K|}\log\left(\frac{2^{K+1}|L|}{\delta}\right)}} &\le
    \frac{\delta}{2^{K+1}|L|}
  \end{align*}
  Pick an arbitrary leaf $A_0$ which is in the tree at time $t_0$.  We can use
  the same approach to find a lower bound on $\hat{s}(A_0)$:
  \begin{align*}
    \P{\hat{s}(A_0) \le s(A_0) -
      \sqrt{\frac{1}{2|A_0|}\log\left(\frac{2^{K+1}|L|}{\delta}\right)}} &\le
    \frac{\delta}{2^{K+1}|L|}
  \end{align*}
  To ensure that $\hat{s}(A_0) \ge \hat{p}(A^K)$ ($\ge \hat{s}(A^K)$) fails to
  hold with probability at most $\delta2^{-K}|L|^{-1}$ we must choose $K$ and
  $t$ to make
  \begin{align*}
    s(A_0) &\ge \mu(A)\left(\frac{2Dm+1}{2Dm+2}\right)^K +
    \sqrt{\frac{1}{2|A^K|}\log\left(\frac{2^{K+1}|L|}{\delta}\right)} +
    \sqrt{\frac{1}{2|A_0|}\log\left(\frac{2^{K+1}|L|}{\delta}\right)}
  \end{align*}
  The first term goes to 0 as $K\to\infty$.  We know that $|A^K| \ge \alpha(K)$
  so the second term also goes to 0 provided that $K/\alpha(K) \to 0$, which we
  require.

  The third term goes to 0 if $K/|A_0| \to 0$.  Recall that $|A_0| =
  N_{t_0,t}^e(A_0)$ and for any $\gamma > 0$
  \begin{align*}
    \P{N_{t_0,t}^e(A) \le N_{t_0,t}^e \mu(A) -
      \sqrt{\frac{1}{2N_{t_0,t}^e}\log\left(\frac{1}{\gamma}\right)}} &\le
    \gamma
  \end{align*}
  From this we see it is sufficient to have $K/N_{t_0,t}^e\to0$ which we
  established in a lemma.

  In summary, there are $|L|$ leafs in the tree at time $t_0$ and each of them
  generates at most $2^K$ different $A^K$'s.  Union bounding over all these
  leafs and over the probability of $N_{t_0,t}^e(A_0)$ growing sublinearly in
  $N_{t_0,t}^e$ we have that, conditioned on the event that $A_0$ has not yet
  been split, $A_0$ is the leaf with the highest value of $\hat{s}$ with
  probability at least $1-\delta-\gamma$ in finite time.  Since $\delta$ and
  $\gamma$ are arbitrary we are done.
\end{proof}


\end{document}